\documentclass[11pt]{article}

\usepackage[final]{acl}

\usepackage{amsmath}
\usepackage{amssymb}
\usepackage{booktabs}
\usepackage{multirow}
\usepackage{algorithm}
\usepackage{algorithmic}
\usepackage{xcolor}
\usepackage{listings}
\usepackage{subcaption}

\usepackage[table]{xcolor}
\usepackage{colortbl}

\usepackage{times}
\usepackage{latexsym}
\usepackage{multicol}

\usepackage[T1]{fontenc}

\usepackage[utf8]{inputenc}

\usepackage{microtype}

\usepackage{inconsolata}

\usepackage{graphicx}

\usepackage{pifont}
\usepackage{mdframed}
\usepackage{enumitem}
\usepackage{needspace}
\usepackage[breakable,skins]{tcolorbox}

\usepackage{siunitx}
\usepackage{listings}

\definecolor{violationred}{RGB}{180,40,40}
\definecolor{repairgreen}{RGB}{225,245,230}
\definecolor{lightgray}{RGB}{245,245,245}

\newcommand{\mytiny}{\fontsize{8.2pt}{7pt}\selectfont}

\usepackage{soul}

\title{Instructions for *ACL Proceedings}

\author{Yuan Tian \\
  Purdue University \\
  \texttt{tian211@purdue.edu} \\\And
  Tianyi Zhang \\
  Purdue University \\
  \texttt{tianyi@purdue.edu} \\}

\definecolor{codegreen}{rgb}{0,0.6,0}
\definecolor{codegray}{rgb}{0.5,0.5,0.5}
\definecolor{codepurple}{rgb}{0.58,0,0.82}
\definecolor{backcolour}{rgb}{0.95,0.95,0.92}

\newcommand{\tool}{\textsc{PV-SQL}}

\newcommand{\metric}[1]{\fontsize{8pt}{5pt}\selectfont \textbf{#1}}

\lstdefinestyle{sqlstyle}{
    backgroundcolor=\color{backcolour},
    commentstyle=\color{codegreen},
    keywordstyle=\color{blue},
    numberstyle=\tiny\color{codegray},
    stringstyle=\color{codepurple},
    basicstyle=\ttfamily\footnotesize,
    breaklines=true,
    frame=single,
}

\title{{\tool}: Synergizing Database Probing and Rule-based Verification for Text-to-SQL Agents}

\begin{document}
\maketitle


\begin{abstract}
Text-to-SQL systems often struggle with deep contextual understanding, particularly for complex queries with subtle requirements.
We present {\tool}, an agentic framework that addresses these failures through two complementary components: \textbf{\underline{P}}robe and \textbf{\underline{V}}erify.
The \textit{Probe} component iteratively generates probing queries to retrieve concrete records from the database, resolving ambiguities in value formats, column semantics, and inter-table relationships to build richer contextual understanding.
The \textit{Verify} component employs a rule-based method to extract verifiable conditions and construct an executable checklist, enabling iterative SQL refinement that effectively reduces missing constraints.
Experiments on the BIRD benchmarks show that {\tool} outperforms the best text-to-SQL baseline by 5\% in execution accuracy and 20.8\% in valid efficiency score while consuming fewer tokens.\footnote{Code is available at \url{https://github.com/magic-YuanTian/PV-SQL}}
\end{abstract}
\section{Introduction}

\begin{figure}[t]
    \centering
    \includegraphics[width=\columnwidth]{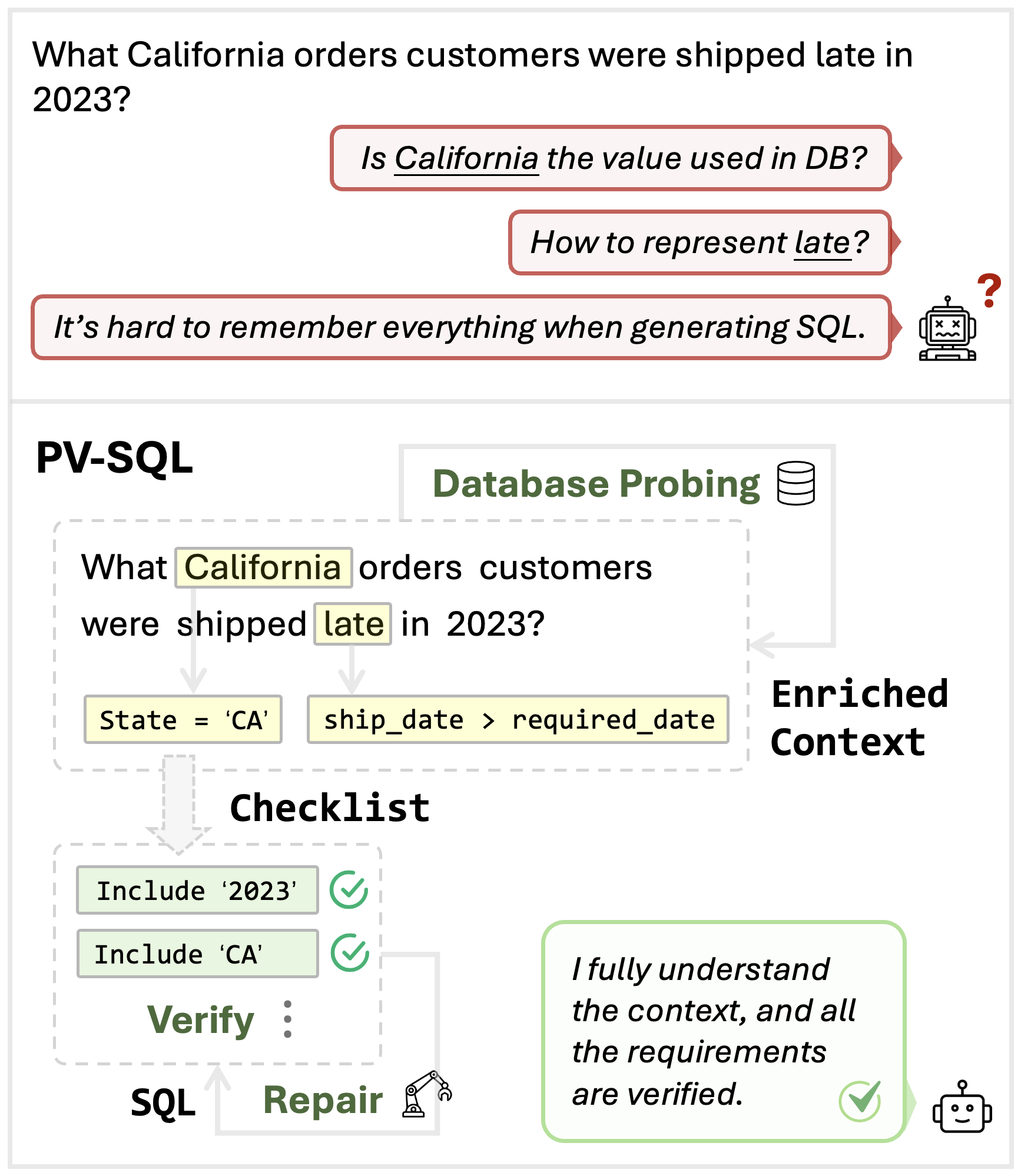}
    \caption{An example of how {\tool} effectively solves a text-to-SQL task.}
    \label{fig:teaser}
\end{figure}

Text-to-SQL has emerged as a critical capability for democratizing database access, enabling non-experts to query structured data using natural language \cite{yu2018spider,li2024bird}. Despite remarkable progress driven by large language models (LLMs), existing systems face persistent challenges: (1) \textbf{schema understanding}, comprehending table relationships and column semantics; (2) \textbf{value grounding}, mapping natural language terms to exact database values, where schema information alone may not help; and (3) \textbf{constraint satisfaction}, ensuring the generated SQL faithfully captures all semantics expressed in the question.

Consider the question (Figure~\ref{fig:teaser}): \textit{``What California orders customers were shipped late in 2023?''} A text-to-SQL system must resolve multiple ambiguities: Is ``California'' stored as the full name or abbreviated as ``CA''? 
How is ``late'' represented? Does the database have a ``late'' flag? Answering these questions requires examining actual database values, yet existing methods typically rely only on schema descriptions, which do not contain such information.\footnote{For example, Data Definition Language (DDL) defines the properties of a database by specifying data types, primary keys, and foreign keys, but does not mention database values.} 
Our empirical study (Section~\ref{sec:study}) reveals that approximately 41\% of failed tasks stem from a misunderstanding of the database.

Previous work mainly uses static methods to enrich the context based on schema linking or semantic enrichment~\cite{li2023resdsql, ren2024purple, e_sql, dcg_sql}. 
However, they only rely on predefined heuristics and do not adaptively explore database content based on the question's needs.
Furthermore, even with a complete understanding of the database context, SQL generation inherently lacks a verification mechanism, such as test cases, to ensure that the generated SQL correctly meets all requirements. Models can easily generate syntactically correct SQL queries that silently return wrong answers, especially for complex queries. 
Although existing work leverages strategies such as refinement, candidate selection, or LLM-based verifications~\cite{askari2024magic, cen2024sqlfixagent, xu2025ts-sql, ni2023lever, sqlens}, a lack of reliable verification mechanisms remains.

To address these challenges, we propose {\tool}, an agentic text-to-SQL method with two complementary components (Figure~\ref{fig:teaser}): (1) \textbf{Database Probing}: Rather than generating SQL based on schema alone, {\tool} can generate and execute probe SQL queries to discover insights from database content, such as value formats, column semantics, and data distributions. (2) \textbf{Verify and Repair}: {\tool} automatically extracts rule-based verifiable constraints from the question using pattern matching (e.g., requiring \texttt{DISTINCT} for ``unique'', \texttt{LIMIT k} for ``top-k''), and iteratively repairs the SQL until all constraints are satisfied.

These two components address complementary failure types. As shown in Figure~\ref{fig:teaser}, \emph{Probe} discovers that ``California'' is stored as ``CA'' and ``late'' means \texttt{ship\_date > required\_date}.
\emph{Verify} then ensures that the SQL includes ``CA'' and ``2023'' as required. Essentially, probing enhances the \emph{input} by enriching the context with concrete evidence, while verification enhances the \emph{output} by ensuring that desired semantic constraints are satisfied.

We evaluate {\tool} by comparing it to seven strong baselines on three evaluation benchmarks across six base LLMs.
{\tool} achieves 65.12\% execution accuracy and 86.9 valid efficiency score on BIRD, outperforming all baselines consistently across all benchmarks. 
Ablation studies confirm that both components contribute significantly.


\section{Related Work}
\label{sec:related_work}

\subsection{Text-to-SQL Methods}
Large language models (LLMs) have transformed text-to-SQL from specialized semantic parsing into a general context understanding and reasoning challenge.
DIN-SQL \cite{pourreza2023dinsql} decomposes the task into sub-problems including schema linking and query classification, while TA-SQL \cite{ta_sql} incorporates task alignment to reduce hallucinations during schema linking and logical synthesis. Some prompt-based frameworks encourage logical checking or multi-path reasoning during generation~\cite{chess,chase_sql}. 
Multi-agent frameworks such as MAC-SQL \cite{wang2023macsql} employ specialized agents for decomposition, generation, and refinement. 
Despite these advances, \citet{rahaman2024sqlunderstanding} show that the latest LLMs struggle with deep semantic understanding, particularly with new schemas~\cite{sqlsynth, evoschema}.

In response to this challenge, {\tool} does not rely on schema alone or free-form self-correction. It enhances LLMs' semantic understanding by (1) actively probing database content to resolve any ambiguity, and (2) rule-based verification to ensure that complex queries do not miss any semantics.

\subsection{Context Enrichment}

A large body of work improves text-to-SQL performance by enriching the model input with additional context, including semantic enrichment, entity linking, schema linking, and information retrieval.
Semantic enrichment methods rewrite or augment the NL query with schema hints or grounded descriptions to make latent requirements explicit.
Entity linking and schema linking methods align mentions in the NL query with database entities to reduce ambiguity.
For example, RESDSQL \cite{li2023resdsql} prioritizes relevant schema elements using ranking-enhanced encoders, while E-SQL~\cite{e_sql} directly enriches questions with linked schema elements and candidate predicates. 
Retrieval-augmented approaches such as PURPLE \cite{ren2024purple} incorporate external demonstrations, but rely on static retrieval based on surface similarity. 
However, these methods mainly identify relevant database entities (e.g., columns), and do not effectively address the value grounding issue by mapping NL terms to database records.

In contrast, {\tool} employs adaptive database probing where the agent iteratively generates temporary SQL queries to explore the database, enabling question-specific context enrichment that similarity-based methods cannot achieve.

\subsection{Verification-Driven Refinement}
Verification-driven methods aim to improve text-to-SQL performance by validating and refining generated queries using feedback signals.

Inspired by self-consistency \cite{wang2022self}, one line of work samples multiple SQL candidates and selects the final output based on comparison or majority voting~ \cite{li2024selectsql, xiyan_sql, sql_of_thought}.
However, these methods are computationally expensive and unreliable, as they require generating many SQL candidates that may contain similar errors.

Another line of research conducts verification using external feedback, such as using the SQL execution results or additional models~\cite{madaan2024selfrefine,askari2024magic,cen2024sqlfixagent, sqlens}.
Self-Refine \cite{madaan2024selfrefine} introduced the idea of using LLMs to critique their own outputs, but \citet{huang2024selfcorrect} show that LLMs struggle to self-correct without external feedback. 
In code generation, test-driven methods such as CodeT \cite{chen2022codet} and Self-Debug \cite{self_debug} leverage generated test cases to refine generation, showing that explicit tests provide stronger verification than LLM self-verification.
However, SQL queries inherently lack explicit test cases. 
LEVER \cite{ni2023lever} checks whether a generated SQL query is runnable but does not consider semantic correctness. 
TS-SQL \cite{xu2025ts-sql} asks the LLM to synthesize test cases based on a translated Python version of the SQL query and then uses test results to refine the SQL.
However, the test cases are still generated by LLMs, which is prone to errors and can lead to error accumulation. Furthermore, these methods often introduce significant computational overhead.
Interactive text-to-SQL generation~\cite{steps, sqlucid} treats the user as the source of feedback signal, soliciting clarifications or corrections to guide refinement. While effective, these methods require a human in the loop and do not scale to automation.

In contrast, {\tool} uses a rule-based verifier that is reliable and lightweight. 
It leverages pattern-matching to extract verifiable constraints from the question, deterministically checks whether the SQL satisfies them, and iteratively repairs the SQL to address violations. 
{\tool} outperforms TS-SQL by 7.8\% execution accuracy on BIRD (Section~\ref{sec:results}).

\section{Problem Statement}
\label{sec:problem}

\subsection{Task Definition}

Given a natural language question $Q$, a database $D$ with schema $O$, and optional evidence $E$ (e.g., additional knowledge of the database), the text-to-SQL task is to generate a SQL query $S$ such that executing it on $D$ returns the correct answer to $Q$:
\begin{equation}
    f: (Q, E, D, O) \rightarrow S
\end{equation}
We decompose the generation process into two stages: 
(1) \emph{understanding}, which builds an internal representation of the question semantics and database content, and 
(2) \emph{synthesis}, which translates this understanding into SQL. 
Formally:
\begin{equation}
    (Q, E, D, O) \xrightarrow{\text{understanding}} R \xrightarrow{\text{synthesis}} S
\end{equation}
where $R$ represents the model's internal reasoning. Errors arise when either stage fails:

\begin{itemize}[leftmargin=*,itemsep=2pt]
    \item \textbf{Database Misinterpretation} ($\mathcal{E}_D$): The understanding $R$ includes incorrect assumptions about database content, e.g., value formats.
    \item \textbf{Question Misinterpretation} ($\mathcal{E}_Q$): The understanding $R$ misses or misinterprets semantic constraints expressed in $Q$.
    \item \textbf{Synthesis Failure} ($\mathcal{E}_S$): The synthesis fails to map a \emph{correct} understanding $R$ into SQL $S$.
\end{itemize}

\subsection{Empirical Study on Error Distribution}
\label{sec:study}

Recent works \cite{ambiSQL, tiis, iui23_empirical, text_to_sql_survey} highlight that ambiguity and context understanding are significant issues in text-to-SQL generation.
\citet{text_to_sql_study} shows that at least 30\% of errors derive from misunderstanding the database schema or the natural language question.
To understand reasons behind LLM-generated text-to-SQL errors, we analyze failed tasks from six LLMs (as discussed in Section~\ref{sec:llm}) on the BIRD benchmark~\cite{li2024bird}. Following previous works~\cite{llm_judge, llm_judge2}, we use GPT-4o as a judge to classify the error type.
We manually measured the LLM judge's correctness in 100 tasks, and the LLM judge achieved 88\% accuracy.
More details are discussed in Appendix~\ref{sec:prompt_error_analysis}.

As shown in Table~\ref{tab:error_analysis}, database misinterpretation ($\mathcal{E}_D$) accounts for 41.3\%, while question misinterpretation ($\mathcal{E}_Q$) accounts for 24.8\%. This suggests that context misunderstanding significantly contributes to generation errors.
This motivates our design: \textbf{Probe} addresses understanding errors ($\mathcal{E}_D + \mathcal{E}_Q$) by grounding the model in actual database content, which directly resolves database misinterpretation and mitigates question misinterpretation that stems from incomplete context. \textbf{Verify} addresses synthesis errors ($\mathcal{E}_S$) by ensuring no semantic constraints from the question are missed.

\begin{table}[t]
\centering
\resizebox{0.85\linewidth}{!}{
\begin{tabular}{lc}
\toprule
\textbf{Error Type} & \textbf{} \\
\midrule
Database Misinterpretation ($\mathcal{E}_D$) & 41.3\% \\
Question Misinterpretation ($\mathcal{E}_Q$) & 24.8\% \\
Synthesis Errors ($\mathcal{E}_S$) & 33.9\% \\
\bottomrule
\end{tabular}
}
\caption{Text-to-SQL Error distribution on BIRD.}
\label{tab:error_analysis}
\end{table}

\section{Method}
\label{sec:method}

Figure~\ref{fig:framework} illustrates {\tool}. 
Given a question $Q$, evidence $E$, and database $D$ with schema $O$, our goal is to generate SQL query $S$ that correctly answers $Q$. 
As motivated in Section~\ref{sec:problem}, we address understanding errors ($\mathcal{E}_D + \mathcal{E}_Q$) by \textbf{Probe} (Figure~\ref{fig:framework}, left) and synthesis errors ($\mathcal{E}_S$) by \textbf{Verify} (Figure~\ref{fig:framework}, right).
Algorithm~\ref{alg:framework} presents the procedure.

\begin{figure*}[t]
    \centering
    \includegraphics[width=\textwidth]{latex/figures/overview.png}
    \caption{Overview of {\tool}. \textbf{Left:} The agent generates probing SQL to discover database content and enriches the context with insights. \textbf{Top:} Rule-based extraction identifies semantic constraints from the question. \textbf{Right:} A rule-based verifier provides feedback for iterative repair until all checks pass.}
    \label{fig:framework}
\end{figure*}

\begin{algorithm}[t]
\caption{{\tool}}
\label{alg:framework}
\begin{algorithmic}[1]
\REQUIRE Question $Q$, Database $D$, Schema $O$, Evidence $E$, Max probes $K$, Max repairs $M$
\ENSURE SQL query $S$
\STATE \textbf{// Database Probing}
\STATE $H \leftarrow \emptyset$ \COMMENT{Probe history}
\STATE $G \leftarrow \emptyset$ \COMMENT{Grounding context}
\FOR{$t = 1$ to $K$}
    \STATE $\text{response} \leftarrow \text{LLM}(Q, E, O, H)$
    \IF{response.action = ``done''}
        \STATE \textbf{break}
    \ENDIF
    \STATE $p_t \leftarrow$ response.probe\_sql
    \STATE $r_t \leftarrow \text{Execute}(D, p_t)$
    \STATE $H \leftarrow H \cup \{(p_t, r_t)\}$
    \STATE Update $G$ using insights from $r_t$
\ENDFOR
\STATE \textbf{// Constraint Extraction}
\STATE $C \leftarrow \text{ExtractConstraints}(Q, E)$
\STATE \textbf{// Verifiable Generation \& Repair}
\STATE $S \leftarrow \text{LLM}(Q, E, G, O, C)$ 
\FOR{$m = 1$ to $M$}
    \STATE $V \leftarrow \text{Violations}(S, C)\ \cup$ \par
    \hspace{\algorithmicindent}\hspace{\algorithmicindent}$\text{ExecutionErrors}(S)$
    \IF{$V = \emptyset$}
        \STATE \textbf{break}
    \ENDIF
    \STATE $S \leftarrow \text{LLM}(Q, E, G, O, C, S, V)$
\ENDFOR
\RETURN $S$
\end{algorithmic}
\end{algorithm}

\subsection{Database Probing}
\label{sec:probing}

The probing component iteratively queries the database to discover value formats and column semantics that the schema alone cannot reveal. Rather than including all database values in the prompt, which is infeasible for large databases and lacks question-specificity, probing enables uncertainty-driven exploration.

\paragraph{Probe Generation.} At each iteration $t$, the LLM receives question, evidence, schema, and probe history. The agent decides whether to issue another probe or proceed to SQL generation. If probing, it generates a \texttt{SELECT} query with a \texttt{LIMIT} clause to retrieve a bounded sample of relevant records.

\paragraph{Context Accumulation.} After each probe execution, the agent interprets the results and summarizes what it has learned, identifying which columns are relevant to the question and extracting value mappings (e.g., ``California'' $\rightarrow$ ``CA''). These insights are accumulated into grounding context $G$, which enriches the subsequent SQL generation with verified database knowledge.
By default, we set the maximum number of probes to $5$.

\subsection{Constraint Extraction}
\label{sec:constraint_extraction}

{\tool} extracts verifiable constraints from the question using pattern-matching rules. This design offers two advantages: (1) \emph{reliability}, as pattern matching is simple and deterministic; (2) \emph{efficiency}, as no additional LLM calls are required. We prioritize precision over recall: missing some constraints is acceptable, but false positives cause unnecessary repairs and introduce errors. We validate this choice empirically in Section~\ref{llm_verify} and Section~\ref{extraction_acc}.

Formally, each rule $r_i \in \mathcal{R}$ maps question patterns to a constraint predicate $c_i$ checkable against SQL. Table~\ref{tab:constraint_rules} summarizes the ten constraint types we support, with representative patterns and their corresponding SQL checks. Complete pattern-matching rules are provided in Appendix~\ref{sec:rules_extract}.

\subsection{Verification and Refinement}
\label{sec:generation}

Given a synthesized SQL query, {\tool} detects potential errors through a multi-step pipeline\footnote{This design enables early termination, where queries failing syntax validation skip database execution entirely, reducing unnecessary computation.}:
\begin{enumerate}[leftmargin=*,itemsep=1pt]
    \item \textbf{Syntax Check}: {\tool} first parses the SQL and checks its syntax correctness without execution using the \texttt{EXPLAIN} command.\footnote{https://www.geeksforgeeks.org/sql/explain-in-sql/}
    \item \textbf{Execution Check}: {\tool} then executes the SQL against the database to catch runtime errors (e.g., type mismatches, division by zero).
    \item \textbf{Constraint Check}: Unlike syntax and execution errors that databases catch automatically, semantic constraint violations require explicit verification. For each extracted constraint $c_i \in C$, we verify the presence of the corresponding SQL construct as specified in Table~\ref{tab:constraint_rules}. Complete verification rules are provided in Appendix~\ref{sec:rules_verify}.
\end{enumerate}

Then, {\tool} repairs constraint-violating SQL by providing the LLM with the failed query, violations, and original context (question, evidence, schema, and grounding context from probing). Each violation produces a descriptive error message that guides targeted repairs. The process terminates when no violations remain, or after a maximum of 5 iterations to prevent an infinite loop.
The verifier is not designed to cover every nuanced or domain-specific constraint; rather, it serves as a high-precision checklist that decomposes complex generation into simpler refining steps, while the LLM remains responsible for capturing constraints outside the rule set during repair.

\begin{table}[h]
\centering
\small
\setlength{\tabcolsep}{2pt}
\begin{tabular}{@{}lll@{}}
\toprule
\textbf{Type} & \textbf{Example NL Patterns} & \textbf{SQL Verification} \\
\midrule
Distinct & unique, distinct, different & \texttt{DISTINCT}, \texttt{GROUP BY} \\
Top-K & top/first/best \textit{N} & \texttt{ORDER BY} + \texttt{LIMIT} \\
Ranking & rank, position, standing & \texttt{RANK()}, \texttt{ROW\_NUMBER()} \\
Count & how many, number of & \texttt{COUNT(*)} \\
Percent & percentage, ratio, rate & division (\texttt{/}) in \texttt{SELECT} \\
Sum & total, sum, combined & \texttt{SUM()} \\
Average & average, mean, avg & \texttt{AVG()} \\
Extreme & max, min, largest, smallest & \texttt{MAX/MIN} or \texttt{LIMIT 1} \\
Temporal & latest, earliest, newest & \texttt{ORDER BY} date column \\
Compare & more/less than, at least & \texttt{>}, \texttt{<}, \texttt{>=}, \texttt{<=} \\
\bottomrule
\end{tabular}
\caption{Simplified constraint extraction and SQL verification rules.}
\label{tab:constraint_rules}
\end{table}



\section{Experimental Setup}

\subsection{Benchmarks}

We evaluate on three widely-used benchmarks. 
First, we use the development set of \textbf{Spider} \cite{yu2018spider}, which includes 1034 tasks. 
Second, we use the development set of a more challenging benchmark, \textbf{BIRD} \cite{li2024bird}, which includes 1534 tasks. Furthermore, NL questions in BIRD are accompanied by evidence or hints that provide domain knowledge. 
Finally, following recent works~\cite{TTD-SQL, rubiksql}, we also use \textbf{Mini-Dev}, a curated 500-example subset of BIRD, developed from community feedback to contain only unambiguous, high-quality queries while preserving the original distribution of difficulty levels and database domains.

\subsection{Metrics}

\paragraph{Execution Accuracy (EX)} measures whether the predicted SQL produces the same result as the gold SQL when executed.

\paragraph{Valid Efficiency Score (VES)} \cite{li2024bird} combines execution accuracy with query efficiency, penalizing slow queries. This is specific to BIRD and Mini-Dev.

\paragraph{Token Consumption.}
Following the insight from test-time scaling~\cite{snell2024scaling}, more reasoning generally leads to better performance. However, token consumption impacts cost and latency in practice. 
We report input tokens, output tokens, and average task completion time per query to evaluate the performance-cost efficiency.

\subsection{Baselines}

We select seven open-source, training-free\footnote{For a fair comparison.} SOTA text-to-SQL methods.
\textbf{DAIL-SQL} \cite{gao2023dailsql} uses example selection and organization so LLMs see useful question–SQL pairs. 
\textbf{DIN-SQL} \cite{pourreza2023dinsql} decomposes the Text-to-SQL task into smaller subproblems and feeds intermediate solutions back to the LLM, boosting reasoning and accuracy compared to naive prompting. 
\textbf{MAC-SQL} \cite{wang2023macsql} is a multi-agent collaborative framework where specialized agents (decomposer, selector, refiner) work together with tools to generate and refine SQL queries for more complex scenarios. 
\textbf{E-SQL} \cite{e_sql} enhances Text-to-SQL by directly enriching the question with relevant schema elements and candidate predicates, improving schema linking and handling ambiguous queries.
\textbf{TA-SQL} \cite{ta_sql} incorporates a task alignment strategy to reduce hallucinations by leveraging experience from similar tasks during schema linking and logical synthesis.
\textbf{XiYan-SQL} \cite{xiyan_sql} generates multiple diverse SQL candidates via a multi-generator ensemble and then selects the best one with a trained selection model.
\textbf{TS-SQL} \cite{xu2025ts-sql} adopts a test-driven refinement method that synthesizes test data and Python scripts that replicate the expected SQL behaviors.These scripts are executed to validate the generated SQL and produce execution feedback, guiding the correction of SQL errors. Since TS-SQL is not publicly available, we replicate it using prompts reported in their paper.

\subsection{Base LLMs}

\label{sec:llm}

Our evaluation includes six base LLMs.
For closed-source models, we use GPT-4o and GPT-4.1-mini. For open-source models, we use GPT-OSS-20B, Gemma3-4B, Qwen3-4B, and Qwen3-0.6B, covering a range of model sizes from 0.6B to 20B parameters. 
All models disable the thinking mode for fair comparison and use a temperature of 0 for reproducibility. 
In Section~\ref{sec:results}, we report results using GPT-4o as the default base model when cross-model evaluation is not the focus.

\begin{table}[h]
\centering
\small

\setlength{\tabcolsep}{3.1pt}
\renewcommand{\arraystretch}{1.15}

\begin{tabular}{l c c p{2pt} c c p{2pt} c}
\toprule
\multirow{2}{*}{\textbf{Method}} &
\multicolumn{2}{c}{\textbf{BIRD}} & &
\multicolumn{2}{c}{\textbf{Mini-Dev}} & &
\multicolumn{1}{c}{\textbf{Spider}} \\
\cmidrule(lr){2-3}\cmidrule(lr){5-6}\cmidrule(lr){8-8}
& \metric{EX} & \metric{VES} &
& \metric{EX} & \metric{VES} &
& \metric{EX} \\
\midrule

GPT-4o
& 52.22 & 56.99 & & 49.40 & 54.57 & & 71.08 \\

\specialrule{0.1pt}{1.3pt}{1.3pt}

+ DAIL-SQL
& 54.10 & 58.77 & & 50.00 & 50.84 & & 71.18 \\

+ DIN-SQL
& 54.80 & 56.75 & & 51.00 & 54.79 & & 69.05 \\

+ MAC-SQL
& 58.70 & 62.77 & & 57.80 & 64.65 & & 72.73 \\

+ E-SQL
& 59.10 & 64.65 & & 57.40 & 64.23 & & 75.63 \\

+ TA-SQL
& 60.43 & 62.99 & & 58.40 & 64.05 & & 74.66 \\

+ XiYan-SQL
& 57.60 & 63.96 & & 52.20 & 56.31 & & 72.73 \\

+ TS-SQL
& 57.37 & 63.43 & & 55.00 & 59.14 & & 74.18 \\

\rowcolor{gray!12}
+ PV-SQL
& \textbf{65.12} & \textbf{75.55} & \cellcolor{gray!12}{}
& \textbf{63.80} & \textbf{74.63} & \cellcolor{gray!12}{}
& \textbf{77.66} \\
\bottomrule
\end{tabular}

\caption{Execution accuracy (EX) and Valid Efficiency Score (VES) of different methods across benchmarks.}
\label{tab:benchmark_cost_tradeoff}
\end{table}

\begin{table*}[h]
\centering
\small

\setlength{\tabcolsep}{4pt}
\renewcommand{\arraystretch}{1.15}

\resizebox{\textwidth}{!}{%
\begin{tabular}{l cc p{2pt} cc p{2pt} cc p{2pt} cc p{2pt} cc p{2pt} cc}

\toprule
\multirow{2}{*}{\textbf{Method}} &
\multicolumn{2}{c}{\textbf{GPT-4o}} & &
\multicolumn{2}{c}{\textbf{GPT-4.1-mini}} & &
\multicolumn{2}{c}{\textbf{GPT-OSS-20B}} & &
\multicolumn{2}{c}{\textbf{Gemma3-4B}} & &
\multicolumn{2}{c}{\textbf{Qwen3-4B}} & &
\multicolumn{2}{c}{\textbf{Qwen3-0.6B}} \\
\cmidrule(lr){2-3}\cmidrule(lr){5-6}\cmidrule(lr){8-9}
\cmidrule(lr){11-12}\cmidrule(lr){14-15}\cmidrule(lr){17-18}
& \metric{EX} & \metric{VES} & 
& \metric{EX} & \metric{VES} & 
& \metric{EX} & \metric{VES} & 
& \metric{EX} & \metric{VES} & 
& \metric{EX} & \metric{VES} & 
& \metric{EX} & \metric{VES} \\
\midrule

ZS-CoT  & 52.22 & 56.99 & & 50.26 & 53.75 & & 50.85 & 52.84 & & 34.55 & 35.65 & & 36.44 & 39.01 & & 13.10 & 15.52 \\
DAIL-SQL   & 54.10 & 58.77 & & 50.40 & 55.05 & & 49.39 & 48.22 & & 27.32 & 27.41 & & 28.81 & 29.99 & & 8.15 & 8.50 \\
DIN-SQL    & 54.80 & 56.75 & & 53.80 & 58.22 & & 50.03 & 46.56 & & 22.23 & 20.57 & & 23.44 & 22.51 & & 6.84 & 7.13 \\
MAC-SQL    & 58.70 & 62.77 & & 55.60 & 60.62 & & 53.59 & 51.50 & & 37.70 & 37.36 & & 39.76 & 40.88 & & 11.60 & 12.95 \\
E-SQL      & 59.10 & 64.65 & & 54.80 & 60.95 & & 53.95 & 53.05 & & 26.20 & 29.14 & & 41.66 & 46.38 & & \textbf{12.84} & \textbf{13.06} \\
TA-SQL     &  60.43 & 62.99 & & 60.20 & 66.11 & & 55.74 & 55.51 & & 23.01 & 22.71 & & 30.20 & 32.48 & & 3.00 & 3.16 \\
XiYan-SQL  & 57.60 & 63.96 & & 51.60 & 58.99 & & 35.98 & 38.03 & & 20.79 & 30.69 & & 36.40 & 38.31 & & 6.60 & 6.93 \\

TS-SQL  & 57.37 & 63.43 & & 53.52 & 58.81 & & 50.98 & 55.20 & & 25.02 & 25.29 & & 27.40 & 31.47 & & 5.60 & 8.48 \\

\rowcolor{gray!12}
PV-SQL
& \textbf{\underline{65.12}} & \textbf{75.55} & \cellcolor{gray!12}{}
& \textbf{63.62} & \textbf{\underline{86.9}} & \cellcolor{gray!12}{}
& \textbf{59.45} & \textbf{61.99} & \cellcolor{gray!12}{}
& \textbf{39.50} & \textbf{42.39} & \cellcolor{gray!12}{}
& \textbf{49.61} & \textbf{51.47} & \cellcolor{gray!12}{}
& 11.08 & 15.39 \\

\bottomrule
\end{tabular}%
}

\caption{Execution accuracy (EX) and Valid Efficiency Score (VES) on the BIRD benchmark across different LLMs and text-to-SQL methods. \underline{Underline} indicates the best performance among all conditions.}
\label{tab:main_results}
\end{table*}

\section{Results}
\label{sec:results}

\subsection{Main Results}

Tables~\ref{tab:main_results} and~\ref{tab:benchmark_cost_tradeoff} present our main results. {\tool} consistently and significantly outperforms all baselines across different LLMs and benchmarks.

\paragraph{Cross-Benchmark Comparison} As shown in Table~\ref{tab:benchmark_cost_tradeoff}, {\tool} consistently outperforms all baselines across the three benchmarks, achieving 65.12\% EX and 75.55\% VES on BIRD, 63.80\% EX and 74.63\% VES on Mini-Dev, and 77.66\% EX on Spider. The results indicate that {\tool} generalizes well across different tasks.

\paragraph{Cross-Model Comparison} As shown in Table~\ref{tab:main_results}, on BIRD, {\tool} achieves 65.12\% EX and 75.55\% VES with GPT-4o, substantially outperforming all baselines by at least 4.69 and 12.56 absolute points, respectively. With GPT-4.1-mini, {\tool} achieves 63.62\% EX and 86.9\% VES, surpassing all baselines by at least 3.42 and 20.79 absolute points. Notably, {\tool} with GPT-4.1-mini achieves a remarkably high VES of 86.9\%.

On open-weight models, {\tool} also demonstrates strong generalizability, achieving the best performance across GPT-OSS-20B (59.45\% EX, 61.99\% VES), Qwen3-4B (49.61\% EX, 51.47\% VES), and Gemma3-4B (39.50\% EX, 42.39\% VES).
However, we observe that agentic methods do not perform well on weaker models, as they have limited ability to process complex agentic workflows~\cite{small_llm_not_good}.
For models smaller than 4B, most baselines reduce performance compared to simple chain-of-thought prompting (ZS-CoT), and Qwen3-0.6B is the only model where {\tool} does not achieve the best performance.

\begin{table}[h]
\centering
\small

\setlength{\tabcolsep}{2.8pt}
\renewcommand{\arraystretch}{1.15}

\begin{tabular}{l c c p{2pt} c c p{2pt} c}
\toprule
\multirow{2}{*}{\textbf{Method}} &
\multicolumn{2}{c}{\textbf{BIRD}} & &
\multicolumn{2}{c}{\textbf{Mini-Dev}} & &
\multicolumn{1}{c}{\textbf{Spider}} \\
\cmidrule(lr){2-3}\cmidrule(lr){5-6}\cmidrule(lr){8-8}
& \metric{EX} & \metric{VES} &
& \metric{EX} & \metric{VES} &
& \metric{EX} \\
\midrule

PV-SQL
& \textbf{65.12} & \textbf{75.55} & & \textbf{63.80} & \textbf{74.63} & & \textbf{77.66} \\

\specialrule{0.1pt}{1.2pt}{1.2pt}

\hspace{0.6em} w/o Probe
& 62.13 & 70.07 & & 60.60 & 69.78 & & 73.79 \\

\hspace{0.6em} w/o Repair
& 61.80 & 74.87 & & 60.80 & 71.07 & & 76.98 \\

\specialrule{0.1pt}{1.2pt}{1.2pt}

\hspace{0.6em} LLM verify
& 59.13 & 65.43 & & 53.80 & 58.27 & & 76.89 \\

\bottomrule
\end{tabular}

\caption{Ablation study of {\tool}.}
\label{tab:ablation_study}
\end{table}

\begin{table}[h]
\centering
\small

\setlength{\tabcolsep}{2.5pt}
\renewcommand{\arraystretch}{1.15}

\begin{tabular}{l cc cc cc}
\toprule
\multirow{2}{*}{\textbf{Method}} &
\multicolumn{2}{c}{\textbf{Simple}}
& \multicolumn{2}{c}{\textbf{Moderate}}
& \multicolumn{2}{c}{\textbf{Challenging}} \\
\cmidrule(lr){2-3}\cmidrule(lr){4-5}\cmidrule(lr){6-7}
& \metric{EX} & \metric{VES}
& \metric{EX} & \metric{VES}
& \metric{EX} & \metric{VES} \\
\midrule

GPT-4o
& 59.35 & 63.34 & 42.89 & 50.56 & 36.55 & 37.07 \\

\specialrule{0.1pt}{1.2pt}{1.2pt}

+ DAIL-SQL
& 63.14 & 70.07 & 42.03 & 43.41 & 35.17 & 35.88 \\
+ DIN-SQL
& 60.97 & 63.30 & 46.12 & 48.57 & 42.76 & 41.09 \\
+ MAC-SQL
& 65.62 & 71.05 & 49.78 & 53.13 & 43.45 & 40.77 \\
+ E-SQL
& 65.08 & 70.31 & 51.29 & 54.67 & 45.52 & 60.54 \\
+ TA-SQL
& 67.57 & 70.38 & 52.16 & 55.05 & 41.38 & 41.22 \\
+ XiYan-SQL
& 64.43 & 71.30 & 48.28 & 54.85 & 44.14 & 46.25 \\
+ TS-SQL
& 63.24 & 63.43 & 50.22 & 52.95 & 42.76 & 44.71 \\

\rowcolor{gray!12}
+ PV-SQL
& \textbf{71.03} & \textbf{81.42} & \textbf{56.68} & \textbf{65.37} & \textbf{54.48} & \textbf{63.51} \\

\bottomrule
\end{tabular}

\caption{Execution accuracy (EX) and Valid Efficiency Score (VES) by task difficulty on BIRD.}
\label{tab:by_difficulty_stacked}
\end{table}

\subsection{Evaluation by Task Difficulties}

Table~\ref{tab:by_difficulty_stacked} breaks down performance by task difficulty on BIRD. 
{\tool} outperforms the best baseline at every difficulty level: 71.03\% vs.~67.57\% (TA-SQL) on Simple, 56.68\% vs.~52.16\% (TA-SQL) on Moderate, and 54.48\% vs.~45.52\% (E-SQL) on Challenging queries. The improvement is most pronounced on \emph{Challenging} (+8.96\% EX over E-SQL), showing that our probe-and-verify approach is particularly effective for complex queries. 
We observe similar patterns on Mini-Dev (Appendix~\ref{sec:appendix_results}).

\subsection{Ablation Study}

We conduct ablation studies to understand the contribution of each component in {\tool}. Table~\ref{tab:ablation_study} shows results when disabling the Probe (``w/o Probe'') or Repair (``w/o Repair'') components, and when replacing our rule-based verification with a LLM-based verification (``LLM Verify'').

\paragraph{Effect of Probing.} Removing database probing leads to performance degradation across all benchmarks, with execution accuracy (EX) dropping by 3.0 points on BIRD, 3.2 points on Mini-Dev, and 3.9 points on Spider. VES also drops substantially by 5.5 points on BIRD, 4.9 points on Mini-Dev, and 3.9 points on Spider. These results confirm that database probing helps the model understand data formats and values.
We provide a detailed ablation study of probing iterations in Appendix~\ref{sec:sensitivity_k}.

\paragraph{Effect of Repair.} The verify-and-repair component is also important. Removing it reduces EX accuracy by 3.3\% on BIRD, 3.0\% on Mini-Dev, and 0.7\% on Spider, indicating that the iterative refinement based on constraint violations is effective.

\paragraph{Rule-based vs.~LLM-based Verification.} 
\label{llm_verify}
We compare our rule-based verification against an LLM-based alternative, where we follow previous works~\cite{self_refine} to design the verification prompts (Appendix~\ref{sec:llm_verify_prompts}).
As shown in Table~\ref{tab:cost_breakdown}, rule-based verification achieves 6\% higher execution accuracy while being 2$\times$ faster and using 45\% fewer tokens, validating our design choice.

\begin{table}[t]
\centering
\resizebox{0.85\linewidth}{!}{
\begin{tabular}{lc}
\toprule
\textbf{Component Evaluation Metrics} & \textbf{} \\
\midrule
Constraint Extraction Acc & 99.39\%\\
\specialrule{0.1pt}{1.3pt}{1.3pt}
Repair Success Rate & 90.82\% \\
Repair Regression Rate & 8.71\% \\
\bottomrule
\end{tabular}
}
\caption{Component evaluation of {\tool} verification and repair. \textit{Constraint Extraction Acc}: gold SQL pass rate. \textit{Repair Success/Regression Rate}: violations resolved / constraints broken during repair.}
\label{tab:component_eval}
\end{table}

\begin{table}[h]
\centering
\small

\setlength{\tabcolsep}{1.8pt}
\renewcommand{\arraystretch}{1.15}

\newcommand{\setcolsize}[1]{%
    \renewcommand{\arraystretch}{1.15}%
    \newcolumntype{s}{>{\fontsize{#1}{#1}\selectfont\centering\arraybackslash}c}%
}

\setcolsize{8.5}  

\begin{tabular}{l
    s s s s
    p{2pt}
    s s s s
}
\toprule
\multirow{2}{*}{\textbf{Method}} &
\multicolumn{4}{c}{\textbf{BIRD}} & &
\multicolumn{4}{c}{\textbf{Mini-Dev}} \\
\cmidrule(lr){2-5}\cmidrule(lr){7-10}
& \multicolumn{1}{c}{\mytiny\textbf{EX}}
& \multicolumn{1}{c}{\mytiny\textbf{In}}
& \multicolumn{1}{c}{\mytiny\textbf{Out}}
& \multicolumn{1}{c}{\mytiny\textbf{Time}}
&
& \multicolumn{1}{c}{\mytiny\textbf{EX}}
& \multicolumn{1}{c}{\mytiny\textbf{In}}
& \multicolumn{1}{c}{\mytiny\textbf{Out}}
& \multicolumn{1}{c}{\mytiny\textbf{Time}} \\
\midrule

ZS-CoT
& 52.2 & 787 & 191 & 1.94 & & 49.4 & 1927 & 211 & 2.1 \\
DAIL-SQL
& 54.1 & 2446 & 50 & 0.8 & & 50.0 & 2247 & 60 & 0.9 \\
DIN-SQL
& 54.8 & 26629 & 629 & 25.0 & & 51.0 & 26630 & 694 & 25.0 \\
MAC-SQL
& 58.7 & 6067 & 387 & 4.7 & & 57.8 & 6203 & 380 & 4.7 \\
E-SQL
& 59.1 & 28177 & 642 & 21.4 & & 57.4 & 29986 & 708 & 24.4 \\
TA-SQL
& 60.43 & 6305 & 242 & 5.3 & & 58.4 & 6365 & 240 & 6.7 \\
XiYan-SQL
& 57.6 & 1689 & 128 & 2.4 & & 52.2 & 1734 & 142 & 1.7 \\
TS-SQL
& 57.4 & 1803 & 310 & 3.6 & & 55.0 & 1891 & 362 & 4.1 \\

\rowcolor{gray!12}
PV-SQL
& 65.1 & 3805 & 248 & 3.4 & \cellcolor{gray!12}{}
& 63.8 & 4270 & 315 & 4.1 \\

- w/o Probe
& 62.1 & 1068 & 67 & 1.0 & & 60.6 & 1191 & 81 & 1.4 \\
- w/o Repair
& 61.8 & 3521 & 223 & 3.1 & & 60.8 & 4261 & 276 & 4.3 \\ 
- LLM verify
& 59.1 & 4887 & 478 & 6.6 & & 53.8 & 4983 & 508 & 6.7 \\

\bottomrule
\end{tabular}

\caption{Efficiency Analysis across benchmarks and baselines. Execution accuracy (EX), input tokens, output tokens, and average task completion time (seconds).}
\label{tab:cost_breakdown}
\end{table}

\subsection{Component Evaluation}
\label{sec:component_eval}

To better understand the performance of individual components, we conduct detailed component evaluation on BIRD. Table~\ref{tab:component_eval} summarizes the results. 


\label{extraction_acc}
\paragraph{Constraint Extraction.} We evaluate constraint extraction accuracy by testing whether the gold SQL satisfies all extracted constraints. If all constraints are satisfied, the extraction is reliable; otherwise, the constraints may be irrelevant and potentially mislead the generation.
Our rule-based method achieves 99.39\% accuracy, suggesting that constraints used in {\tool} are highly reliable. 

\paragraph{Repair.} We measure the repair performance by \emph{Repair Success Rate} (violations successfully addressed) and \emph{Regression Rate} (previously satisfied constraints broken). {\tool} achieves a 90.82\% success rate with only 8.71\% regression rate, showing that repairs reliably correct errors without introducing new errors in most cases. 






\begin{figure}[h]
    \centering
    \includegraphics[width=\columnwidth]{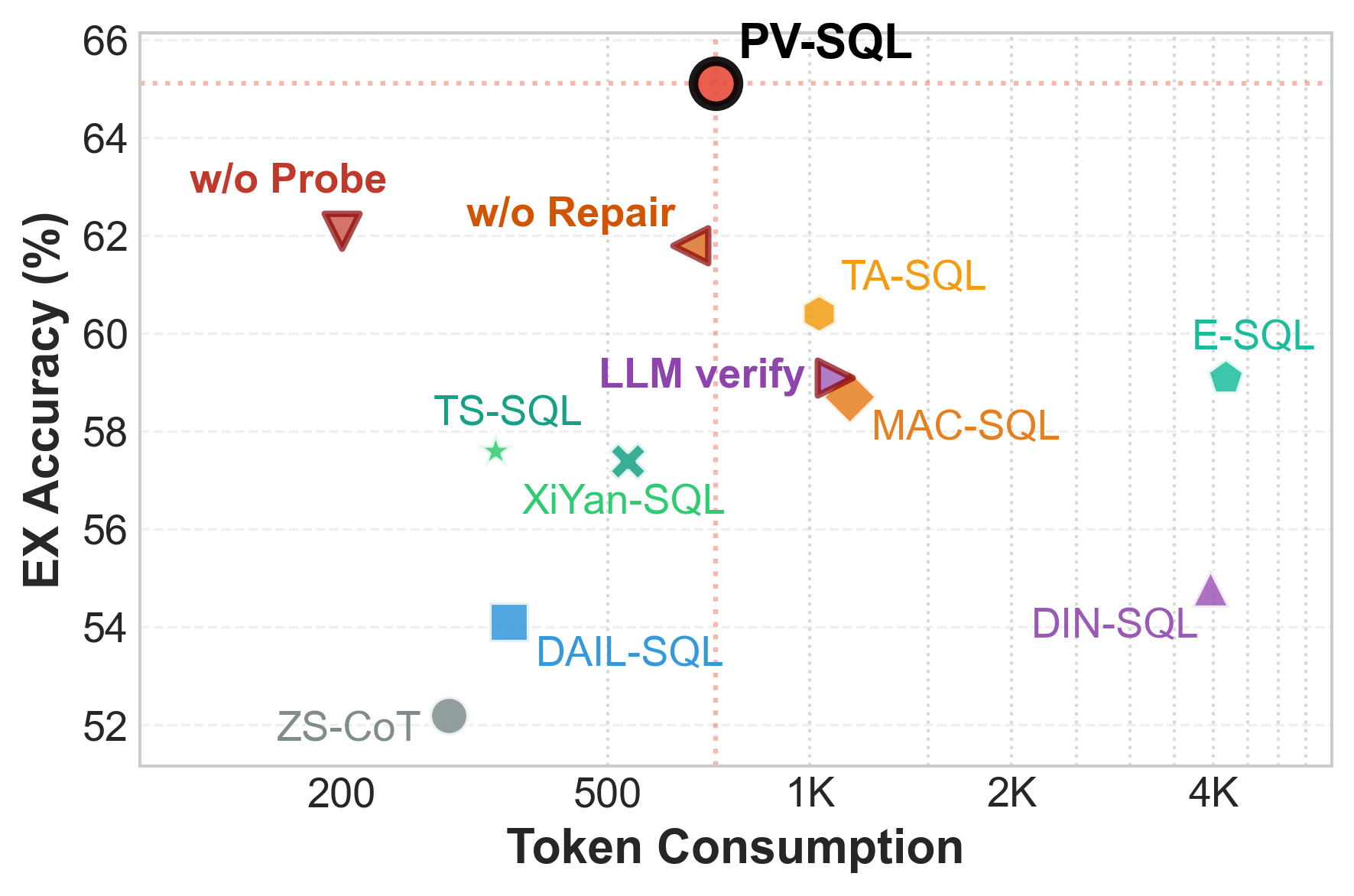}
    \caption{Accuracy vs.\ token consumption\protect\footnotemark{} on BIRD.}
    \label{fig:ex_vs_token}
\end{figure}
\footnotetext{Token consumption is computed as $\frac{1}{8}\times$input + output tokens, reflecting API pricing where input tokens cost $\sim$8$\times$ less than output (\url{https://openai.com/api/pricing/}).}

\subsection{Efficiency Analysis}

Table~\ref{tab:cost_breakdown} reports detailed efficiency metrics (token consumption, latency), while Figure~\ref{fig:ex_vs_token} visualizes the accuracy-efficiency landscape. {\tool} achieves the best accuracy (65.1\% on BIRD, 63.8\% on Mini-Dev) with moderate token consumption. In contrast, DIN-SQL and E-SQL consume 7--8$\times$ more tokens and take 6--7$\times$ longer, yet achieve lower accuracy. The efficiency stems from our rule-based verification, which avoids expensive LLM calls. Replacing it with ``LLM verify'' increases tokens significantly while decreasing accuracy by 6\%. 
Notably, removing Probe achieves the best efficiency with only a 3\% drop in accuracy.

\subsection{Error Analysis}

\begin{figure}[t]
    \centering
    \includegraphics[width=\columnwidth]{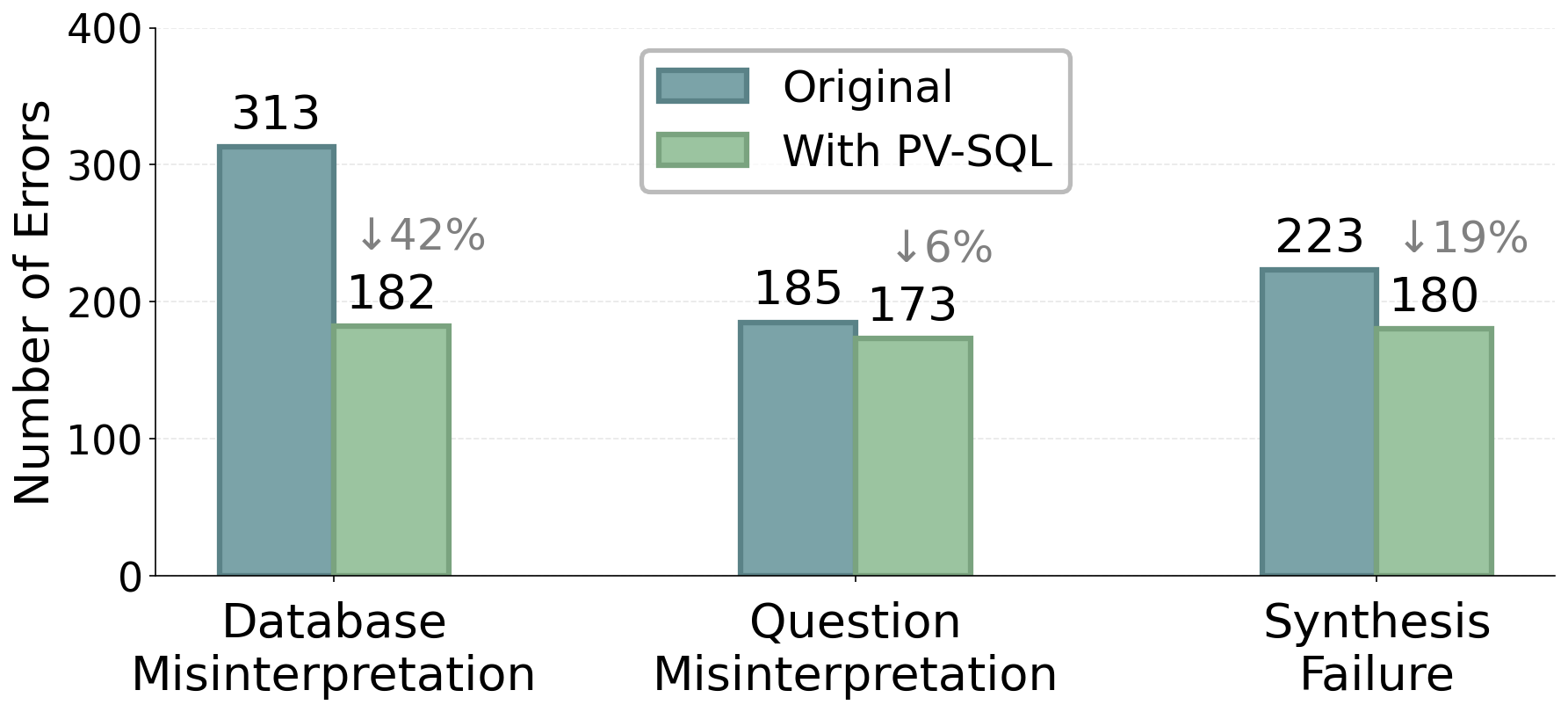}
    \caption{Error distribution w/ and w/o {\tool}. }
    \label{fig:error_reduction}
\end{figure}

To understand how {\tool} reduces errors, we apply the same error classification method as in Section~\ref{sec:problem} to the BIRD benchmark.
Figure~\ref{fig:error_reduction} shows the error distribution before and after applying {\tool}. 
Specifically, there is a 42\% reduction in database misinterpretation errors, validating our hypothesis that probing helps ground the model in actual database content. 
Synthesis failures also decrease by 19\%, demonstrating that the verify-and-repair loop effectively catches semantic constraint violations. 
Question misinterpretation shows the smallest reduction (6\%), which is expected since this error type often stems from inherent ambiguity in natural language rather than missing context.

\section{Conclusion}

We presented {\tool}, an agentic method for text-to-SQL generation. By probing the database to enrich the input and extracting constraints to verify the output, {\tool} addresses complementary errors and consistently enhances text-to-SQL performance with lower token consumption.

\section*{Limitations}

Our constraint extraction rules were developed through empirical observation of common text-to-SQL patterns. While the current rule set covers frequently occurring constraints (e.g., aggregation, sorting, filtering), it does not exhaustively capture all possible semantic requirements. We deliberately favor precision over recall so that the verifier acts as a reliable checklist (99.39\% accuracy, Table~\ref{tab:component_eval}) and leave nuanced or domain-specific constraints to be handled by the LLM during refinement; an analysis of the residual headroom from missed constraints is given in Appendix~\ref{sec:constraint_coverage}. Extending the rule set to handle more such constraints remains an avenue for future work.

Additionally, our empirical error analysis for text-to-SQL relies on LLM-as-a-judge.
Although our human annotation suggests that this setup is reasonably reliable and scalable, it may still introduce inaccuracies.
The automated assessments should be interpreted as estimates rather than ground truth.


\bibliography{custom}

@inproceedings{yu2018spider,
    title = "{S}pider: A Large-Scale Human-Labeled Dataset for Complex and Cross-Domain Semantic Parsing and Text-to-{SQL} Task",
    author = "Yu, Tao and Zhang, Rui and Yang, Kai and Yasunaga, Michihiro and Wang, Dongxu and Li, Zifan and Ma, James and Li, Irene and Yao, Qingning and Roman, Shanelle and Zhang, Zilin and Radev, Dragomir",
    booktitle = "Proceedings of the 2018 Conference on Empirical Methods in Natural Language Processing",
    month = oct,
    year = "2018",
    address = "Brussels, Belgium",
    publisher = "Association for Computational Linguistics",
    pages = "3911--3921"
}

@inproceedings{li2024bird,
    title = "Can {LLM} Already Serve as A Database Interface? A {B}ig Bench for Large-Scale Database Grounded Text-to-{SQL}s",
    author = "Li, Jinyang and Hui, Binyuan and Qu, Ge and Yang, Jiaxi and Li, Binhua and Li, Bowen and Wang, Bailin and Qin, Bowen and Geng, Ruiying and Huo, Nan and Zhou, Xuanhe and Ma, Chenhao and Li, Guoliang and Chang, Kevin Chen-Chuan and Li, Fei and Hui, Bei and Li, Yongbin",
    booktitle = "Advances in Neural Information Processing Systems",
    volume = "36",
    year = "2024"
}

@misc{e_sql,
      title={E-SQL: Direct Schema Linking via Question Enrichment in Text-to-SQL}, 
      author={Hasan Alp Caferoğlu and Özgür Ulusoy},
      year={2025},
      eprint={2409.16751},
      archivePrefix={arXiv},
      primaryClass={cs.CL},
      url={https://arxiv.org/abs/2409.16751}, 
}

@inproceedings{li2023resdsql,
author = {Li, Haoyang and Zhang, Jing and Li, Cuiping and Chen, Hong},
title = {RESDSQL: decoupling schema linking and skeleton parsing for text-to-SQL},
year = {2023},
isbn = {978-1-57735-880-0},
publisher = {AAAI Press},
url = {https://doi.org/10.1609/aaai.v37i11.26535},
doi = {10.1609/aaai.v37i11.26535},
booktitle = {Proceedings of the Thirty-Seventh AAAI Conference on Artificial Intelligence and Thirty-Fifth Conference on Innovative Applications of Artificial Intelligence and Thirteenth Symposium on Educational Advances in Artificial Intelligence},
articleno = {1466},
numpages = {9},
series = {AAAI'23/IAAI'23/EAAI'23}
}

@article{wang2022self,
  title={Self-Consistency Improves Chain of Thought Reasoning in Language Models},
  author={Wang, Xuezhi and Wei, Jason and Schuurmans, Dale and Le, Quoc and Chi, Ed and Narang, Sharan and Chowdhery, Aakanksha and Zhou, Denny},
  journal={arXiv preprint arXiv:2203.11171},
  year={2022}
}

@misc{self_refine,
      title={Self-Refine: Iterative Refinement with Self-Feedback}, 
      author={Aman Madaan and Niket Tandon and Prakhar Gupta and Skyler Hallinan and Luyu Gao and Sarah Wiegreffe and Uri Alon and Nouha Dziri and Shrimai Prabhumoye and Yiming Yang and Shashank Gupta and Bodhisattwa Prasad Majumder and Katherine Hermann and Sean Welleck and Amir Yazdanbakhsh and Peter Clark},
      year={2023},
      eprint={2303.17651},
      archivePrefix={arXiv},
      primaryClass={cs.CL},
      url={https://arxiv.org/abs/2303.17651}, 
}

@misc{self_debug,
      title={Teaching Large Language Models to Self-Debug}, 
      author={Xinyun Chen and Maxwell Lin and Nathanael Schärli and Denny Zhou},
      year={2023},
      eprint={2304.05128},
      archivePrefix={arXiv},
      primaryClass={cs.CL},
      url={https://arxiv.org/abs/2304.05128}, 
}

@inproceedings{ni2023lever,
author = {Ni, Ansong and Iyer, Srini and Radev, Dragomir and Stoyanov, Ves and Yih, Wen-tau and Wang, Sida I. and Lin, Xi Victoria},
title = {LEVER: learning to verify language-to-code generation with execution},
year = {2023},
publisher = {JMLR.org},
booktitle = {Proceedings of the 40th International Conference on Machine Learning},
articleno = {1086},
numpages = {23},
location = {Honolulu, Hawaii, USA},
series = {ICML'23}
}

@misc{snell2024scaling,
      title={Scaling LLM Test-Time Compute Optimally can be More Effective than Scaling Model Parameters}, 
      author={Charlie Snell and Jaehoon Lee and Kelvin Xu and Aviral Kumar},
      year={2024},
      eprint={2408.03314},
      archivePrefix={arXiv},
      primaryClass={cs.LG},
      url={https://arxiv.org/abs/2408.03314}, 
}

@inproceedings{gao2023dailsql,
    title = "{DAIL}-{SQL}: Optimized LLM Prompt for Text-to-{SQL}",
    author = "Gao, Dawei and Wang, Haibin and Li, Yaliang and Sun, Xiuyu and Qian, Yichen and Ding, Bolin and Zhou, Jingren",
    booktitle = "Proceedings of the VLDB Endowment",
    volume = "17",
    number = "5",
    pages = "950--961",
    year = "2024"
}

@misc{pourreza2023dinsql,
    title={DIN-SQL: Decomposed In-Context Learning of Text-to-SQL with Self-Correction},
    author={Mohammadreza Pourreza and Davood Rafiei},
    year={2023},
    eprint={2304.11015},
    archivePrefix={arXiv},
    primaryClass={cs.CL}
}

@misc{wang2023macsql,
    title={MAC-SQL: A Multi-Agent Collaborative Framework for Text-to-SQL},
    author={Bing Wang and Changyu Ren and Jian Yang and Xinnian Liang and Jiaqi Bai and Linzheng Chai and Zhao Yan and Qian-Wen Zhang and Di Yin and Xing Sun and Zhoujun Li},
    year={2024},
    eprint={2312.11242},
    archivePrefix={arXiv},
    primaryClass={cs.CL}
}

@misc{rahaman2024sqlunderstanding,
    title={Evaluating SQL Understanding in Large Language Models},
    author={Md Mahadi Hassan Rahaman and Mehmet Emre Gursoy},
    year={2024},
    eprint={2410.10680},
    archivePrefix={arXiv},
    primaryClass={cs.DB}
}

@misc{ren2024purple,
    title={PURPLE: Making a Large Language Model a Better SQL Writer},
    author={Minseok Ren and Sijie Jin and Soobin Son and Yeonsu Kwon and Dong-Gun Lee and Hwi-Jun Yang and Kyomin Jung},
    year={2024},
    eprint={2403.20014},
    archivePrefix={arXiv},
    primaryClass={cs.CL}
}

@misc{huang2024selfcorrect,
    title={Large Language Models Cannot Self-Correct Reasoning Yet},
    author={Jie Huang and Xinyun Chen and Swaroop Mishra and Huaixiu Steven Zheng and Adams Wei Yu and Xinying Song and Denny Zhou},
    year={2024},
    eprint={2310.01798},
    archivePrefix={arXiv},
    primaryClass={cs.CL}
}

@misc{chen2022codet,
    title={CodeT: Code Generation with Generated Tests},
    author={Bei Chen and Fengji Zhang and Anh Nguyen and Daoguang Zan and Zeqi Lin and Jian-Guang Lou and Weizhu Chen},
    year={2022},
    eprint={2207.10397},
    archivePrefix={arXiv},
    primaryClass={cs.SE}
}

@inproceedings{xu2025ts-sql,
  title={{TS-SQL}: Test-driven Self-refinement for Text-to-{SQL}},
  author={Xu, Wenbo and Zhu, Haifeng and Yan, Liang and Liu, Chuanyi and Han, Peiyi and Duan, Shaoming and Pan, Jeff Z.},
  booktitle={Findings of the Association for Computational Linguistics: EMNLP 2025},
  pages={2864--2889},
  year={2025}
}

@article{askari2024magic,
  title={{MAGIC}: Generating Self-Correction Guidelines for In-Context Text-to-{SQL}},
  author={Askari, Arian and Poelitz, Christian and Tang, Xinye},
  journal={arXiv preprint arXiv:2406.12692},
  year={2024}
}

@article{cen2024sqlfixagent,
  title={{SQLFixAgent}: Towards Semantic-Accurate {SQL} Generation via Multi-Agent Collaboration},
  author={Cen, Jipeng and Liu, Jiaxin and Li, Zhixu and Wang, Jingjing},
  journal={arXiv preprint arXiv:2406.13408},
  year={2024}
}

@article{li2024selectsql,
  title={Using {LLM} to Select the Right {SQL} Query from Candidates},
  author={Li, Zhenwen and Xie, Tao},
  journal={arXiv preprint arXiv:2401.02115},
  year={2024}
}

@article{madaan2024selfrefine,
  title={Self-Refine: Iterative Refinement with Self-Feedback},
  author={Madaan, Aman and Tandon, Niket and Gupta, Prakhar and Hallinan, Skyler and Gao, Luyu and Wiegreffe, Sarah and Alon, Uri and Dziri, Nouha and Prabhumoye, Shrimai and Yang, Yiming and others},
  journal={Advances in Neural Information Processing Systems},
  volume={36},
  year={2024}
}

@misc{sql_of_thought,
      title={SQL-of-Thought: Multi-agentic Text-to-SQL with Guided Error Correction}, 
      author={Saumya Chaturvedi and Aman Chadha and Laurent Bindschaedler},
      year={2025},
      eprint={2509.00581},
      archivePrefix={arXiv},
      primaryClass={cs.DB},
      url={https://arxiv.org/abs/2509.00581}, 
}

@misc{ta_sql,
      title={Before Generation, Align it! A Novel and Effective Strategy for Mitigating Hallucinations in Text-to-SQL Generation}, 
      author={Ge Qu and Jinyang Li and Bowen Li and Bowen Qin and Nan Huo and Chenhao Ma and Reynold Cheng},
      year={2024},
      eprint={2405.15307},
      archivePrefix={arXiv},
      primaryClass={cs.CL},
      url={https://arxiv.org/abs/2405.15307}, 
}

@misc{chase_sql,
      title={CHASE-SQL: Multi-Path Reasoning and Preference Optimized Candidate Selection in Text-to-SQL}, 
      author={Mohammadreza Pourreza and Hailong Li and Ruoxi Sun and Yeounoh Chung and Shayan Talaei and Gaurav Tarlok Kakkar and Yu Gan and Amin Saberi and Fatma Ozcan and Sercan O. Arik},
      year={2024},
      eprint={2410.01943},
      archivePrefix={arXiv},
      primaryClass={cs.LG},
      url={https://arxiv.org/abs/2410.01943}, 
}

@misc{chess,
      title={CHESS: Contextual Harnessing for Efficient SQL Synthesis}, 
      author={Shayan Talaei and Mohammadreza Pourreza and Yu-Chen Chang and Azalia Mirhoseini and Amin Saberi},
      year={2024},
      eprint={2405.16755},
      archivePrefix={arXiv},
      primaryClass={cs.LG},
      url={https://arxiv.org/abs/2405.16755}, 
}

@misc{xiyan_sql,
      title={XiYan-SQL: A Novel Multi-Generator Framework For Text-to-SQL}, 
      author={Yifu Liu and Yin Zhu and Yingqi Gao and Zhiling Luo and Xiaoxia Li and Xiaorong Shi and Yuntao Hong and Jinyang Gao and Yu Li and Bolin Ding and Jingren Zhou},
      year={2025},
      eprint={2507.04701},
      archivePrefix={arXiv},
      primaryClass={cs.CL},
      url={https://arxiv.org/abs/2507.04701}, 
}

@misc{ambiSQL,
      title={AmbiSQL: Interactive Ambiguity Detection and Resolution for Text-to-SQL}, 
      author={Zhongjun Ding and Yin Lin and Tianjing Zeng},
      year={2025},
      eprint={2508.15276},
      archivePrefix={arXiv},
      primaryClass={cs.DB},
      url={https://arxiv.org/abs/2508.15276}, 
}

@misc{text_to_sql_survey,
      title={A Survey of Text-to-SQL in the Era of LLMs: Where are we, and where are we going?}, 
      author={Xinyu Liu and Shuyu Shen and Boyan Li and Peixian Ma and Runzhi Jiang and Yuxin Zhang and Ju Fan and Guoliang Li and Nan Tang and Yuyu Luo},
      year={2025},
      eprint={2408.05109},
      archivePrefix={arXiv},
      primaryClass={cs.DB},
      url={https://arxiv.org/abs/2408.05109}, 
}

@misc{sqlens,
      title={SQLens: An End-to-End Framework for Error Detection and Correction in Text-to-SQL}, 
      author={Yue Gong and Chuan Lei and Xiao Qin and Kapil Vaidya and Balakrishnan Narayanaswamy and Tim Kraska},
      year={2025},
      eprint={2506.04494},
      archivePrefix={arXiv},
      primaryClass={cs.CL},
      url={https://arxiv.org/abs/2506.04494}, 
}

@misc{text_to_sql_study,
      title={A Study of In-Context-Learning-Based Text-to-SQL Errors}, 
      author={Jiawei Shen and Chengcheng Wan and Ruoyi Qiao and Jiazhen Zou and Hang Xu and Yuchen Shao and Yueling Zhang and Weikai Miao and Geguang Pu},
      year={2025},
      eprint={2501.09310},
      archivePrefix={arXiv},
      primaryClass={cs.CL},
      url={https://arxiv.org/abs/2501.09310}, 
}

@misc{rubiksql,
      title={RubikSQL: Lifelong Learning Agentic Knowledge Base as an Industrial NL2SQL System}, 
      author={Zui Chen and Han Li and Xinhao Zhang and Xiaoyu Chen and Chunyin Dong and Yifeng Wang and Xin Cai and Su Zhang and Ziqi Li and Chi Ding and Jinxu Li and Shuai Wang and Dousheng Zhao and Sanhai Gao and Guangyi Liu},
      year={2025},
      eprint={2508.17590},
      archivePrefix={arXiv},
      primaryClass={cs.DB},
      url={https://arxiv.org/abs/2508.17590}, 
}

@inproceedings{TTD-SQL,
    title = "{TTD}-{SQL}: Tree-Guided Token Decoding for Efficient and Schema-Aware {SQL} Generation",
    author = "Sharma, Chetan  and
      Narayanam, Ramasuri  and
      Pal, Soumyabrata  and
      Yeturu, Kalidas  and
      Saini, Shiv Kumar  and
      Mukherjee, Koyel",
    editor = "Potdar, Saloni  and
      Rojas-Barahona, Lina  and
      Montella, Sebastien",
    booktitle = "Proceedings of the 2025 Conference on Empirical Methods in Natural Language Processing: Industry Track",
    month = nov,
    year = "2025",
    address = "Suzhou (China)",
    publisher = "Association for Computational Linguistics",
    url = "https://aclanthology.org/2025.emnlp-industry.90/",
    doi = "10.18653/v1/2025.emnlp-industry.90",
    pages = "1287--1298",
    ISBN = "979-8-89176-333-3"
}

@inproceedings{dcg_sql,
    title = "{DCG}-{SQL}: Enhancing In-Context Learning for Text-to-{SQL} with Deep Contextual Schema Link Graph",
    author = "Lee, Jihyung  and
      Lee, Jin-Seop  and
      Lee, Jaehoon  and
      Choi, YunSeok  and
      Lee, Jee-Hyong",
    editor = "Che, Wanxiang  and
      Nabende, Joyce  and
      Shutova, Ekaterina  and
      Pilehvar, Mohammad Taher",
    booktitle = "Proceedings of the 63rd Annual Meeting of the Association for Computational Linguistics (Volume 1: Long Papers)",
    month = jul,
    year = "2025",
    address = "Vienna, Austria",
    publisher = "Association for Computational Linguistics",
    url = "https://aclanthology.org/2025.acl-long.748/",
    doi = "10.18653/v1/2025.acl-long.748",
    pages = "15397--15412",
    ISBN = "979-8-89176-251-0"
}

@inproceedings{small_llm_not_good,
    title = "Small {LLM}s Are Weak Tool Learners: A Multi-{LLM} Agent",
    author = "Shen, Weizhou  and
      Li, Chenliang  and
      Chen, Hongzhan  and
      Yan, Ming  and
      Quan, Xiaojun  and
      Chen, Hehong  and
      Zhang, Ji  and
      Huang, Fei",
    editor = "Al-Onaizan, Yaser  and
      Bansal, Mohit  and
      Chen, Yun-Nung",
    booktitle = "Proceedings of the 2024 Conference on Empirical Methods in Natural Language Processing",
    month = nov,
    year = "2024",
    address = "Miami, Florida, USA",
    publisher = "Association for Computational Linguistics",
    url = "https://aclanthology.org/2024.emnlp-main.929/",
    doi = "10.18653/v1/2024.emnlp-main.929",
    pages = "16658--16680"
}

@inproceedings{llm_judge,
author = {Zheng, Lianmin and Chiang, Wei-Lin and Sheng, Ying and Zhuang, Siyuan and Wu, Zhanghao and Zhuang, Yonghao and Lin, Zi and Li, Zhuohan and Li, Dacheng and Xing, Eric P. and Zhang, Hao and Gonzalez, Joseph E. and Stoica, Ion},
title = {Judging LLM-as-a-judge with MT-bench and Chatbot Arena},
year = {2023},
publisher = {Curran Associates Inc.},
address = {Red Hook, NY, USA},
booktitle = {Proceedings of the 37th International Conference on Neural Information Processing Systems},
articleno = {2020},
numpages = {29},
location = {New Orleans, LA, USA},
series = {NIPS '23}
}

@misc{llm_judge2,
      title={LLM-as-a-qualitative-judge: automating error analysis in natural language generation}, 
      author={Nadezhda Chirkova and Tunde Oluwaseyi Ajayi and Seth Aycock and Zain Muhammad Mujahid and Vladana Perlić and Ekaterina Borisova and Markarit Vartampetian},
      year={2025},
      eprint={2506.09147},
      archivePrefix={arXiv},
      primaryClass={cs.CL},
      url={https://arxiv.org/abs/2506.09147}, 
}

@inproceedings{sqlucid,
author = {Tian, Yuan and Kummerfeld, Jonathan K. and Li, Toby Jia-Jun and Zhang, Tianyi},
title = {SQLucid: Grounding Natural Language Database Queries with Interactive Explanations},
year = {2024},
isbn = {9798400706288},
publisher = {Association for Computing Machinery},
address = {New York, NY, USA},
url = {https://doi.org/10.1145/3654777.3676368},
doi = {10.1145/3654777.3676368},
booktitle = {Proceedings of the 37th Annual ACM Symposium on User Interface Software and Technology},
articleno = {12},
numpages = {20},
location = {Pittsburgh, PA, USA},
series = {UIST '24}
}

@inproceedings{steps,
    title = "Interactive Text-to-{SQL} Generation via Editable Step-by-Step Explanations",
    author = "Tian, Yuan  and
      Zhang, Zheng  and
      Ning, Zheng  and
      Li, Toby Jia-Jun  and
      Kummerfeld, Jonathan K.  and
      Zhang, Tianyi",
    editor = "Bouamor, Houda  and
      Pino, Juan  and
      Bali, Kalika",
    booktitle = "Proceedings of the 2023 Conference on Empirical Methods in Natural Language Processing",
    month = dec,
    year = "2023",
    address = "Singapore",
    publisher = "Association for Computational Linguistics",
    url = "https://aclanthology.org/2023.emnlp-main.1004/",
    doi = "10.18653/v1/2023.emnlp-main.1004",
    pages = "16149--16166"
}

@inproceedings{sqlsynth,
author = {Tian, Yuan and Lee, Daniel and Wu, Fei and Mai, Tung and Qian, Kun and Sahai, Siddhartha and Zhang, Tianyi and Li, Yunyao},
title = {Text-to-SQL Domain Adaptation via Human-LLM Collaborative Data Annotation},
year = {2025},
isbn = {9798400713064},
publisher = {Association for Computing Machinery},
address = {New York, NY, USA},
url = {https://doi.org/10.1145/3708359.3712083},
doi = {10.1145/3708359.3712083},
booktitle = {Proceedings of the 30th International Conference on Intelligent User Interfaces},
pages = {1398–1425},
numpages = {28},
location = {
},
series = {IUI '25}
}

@article{tiis,
author = {Ning, Zheng and Tian, Yuan and Zhang, Zheng and Zhang, Tianyi and Li, Toby Jia-Jun},
title = {Insights into Natural Language Database Query Errors: from Attention Misalignment to User Handling Strategies},
year = {2024},
issue_date = {December 2024},
publisher = {Association for Computing Machinery},
address = {New York, NY, USA},
volume = {14},
number = {4},
issn = {2160-6455},
url = {https://doi.org/10.1145/3650114},
doi = {10.1145/3650114},
journal = {ACM Trans. Interact. Intell. Syst.},
month = dec,
articleno = {25},
numpages = {32}
}

@inproceedings{iui23_empirical,
author = {Ning, Zheng and Zhang, Zheng and Sun, Tianyi and Tian, Yuan and Zhang, Tianyi and Li, Toby Jia-Jun},
title = {An Empirical Study of Model Errors and User Error Discovery and Repair Strategies in Natural Language Database Queries},
year = {2023},
isbn = {9798400701061},
publisher = {Association for Computing Machinery},
address = {New York, NY, USA},
url = {https://doi.org/10.1145/3581641.3584067},
doi = {10.1145/3581641.3584067},
booktitle = {Proceedings of the 28th International Conference on Intelligent User Interfaces},
pages = {633–649},
numpages = {17},
location = {Sydney, NSW, Australia},
series = {IUI '23}
}

@article{evoschema,
author = {Zhang, Tianshu and Qian, Kun and Sahai, Siddhartha and Tian, Yuan and Garg, Shaddy and Sun, Huan and Li, Yunyao},
title = {Evoschema: Towards Text-to-SQL Robustness against Schema Evolution},
year = {2025},
issue_date = {June 2025},
publisher = {VLDB Endowment},
volume = {18},
number = {10},
issn = {2150-8097},
url = {https://doi.org/10.14778/3748191.3748222},
doi = {10.14778/3748191.3748222},
journal = {Proc. VLDB Endow.},
month = jun,
pages = {3655–3668},
numpages = {14}
}

@misc{all_fem,
      title={ALL-FEM: Agentic Large Language models Fine-tuned for Finite Element Methods}, 
      author={Rushikesh Deotale and Adithya Srinivasan and Yuan Tian and Tianyi Zhang and Pavlos Vlachos and Hector Gomez},
      year={2026},
      eprint={2603.21011},
      archivePrefix={arXiv},
      primaryClass={cs.CE},
      url={https://arxiv.org/abs/2603.21011}, 
}

@misc{alloy,
      title={ALLOY: Generating Reusable Agent Workflows from User Demonstration}, 
      author={Jiawen Li and Zheng Ning and Yuan Tian and Toby Jia-jun Li},
      year={2025},
      eprint={2510.10049},
      archivePrefix={arXiv},
      primaryClass={cs.HC},
      url={https://arxiv.org/abs/2510.10049}, 
}

@misc{atar,
      title={Attention-Aligned Reasoning for Large Language Models}, 
      author={Hongxiang Zhang and Yuan Tian and Tianyi Zhang},
      year={2026},
      eprint={2510.03223},
      archivePrefix={arXiv},
      primaryClass={cs.CL},
      url={https://arxiv.org/abs/2510.03223}, 
}

@misc{spa,
      title={Selective Prompt Anchoring for Code Generation}, 
      author={Yuan Tian and Tianyi Zhang},
      year={2025},
      eprint={2408.09121},
      archivePrefix={arXiv},
      primaryClass={cs.LG},
      url={https://arxiv.org/abs/2408.09121}, 
}

@misc{construct_ai,
      title={Supporting Construction Worker Well-Being with a Multi-Agent Conversational AI System}, 
      author={Fan Yang and Yuan Tian and Jiansong Zhang},
      year={2025},
      eprint={2506.07997},
      archivePrefix={arXiv},
      primaryClass={cs.HC},
      url={https://arxiv.org/abs/2506.07997}, 
}

\newpage

\appendix

\definecolor{promptbg}{RGB}{248,248,248}
\definecolor{promptframe}{RGB}{200,200,200}
\definecolor{sectioncolor}{RGB}{0,0,139}

\newenvironment{prompt}[1]{%
    \begin{figure*}[t]
    \noindent\textbf{#1}
    \vspace{0.3em}
    \begin{tcolorbox}[
        breakable=false,
        enhanced,
        colback=promptbg,
        colframe=promptframe,
        boxrule=0.5pt,
        arc=3pt,
        left=6pt,
        right=6pt,
        top=6pt,
        bottom=6pt,
        width=\textwidth,
        fontupper=\small\ttfamily\raggedright,
        parbox=false
    ]
    \begin{multicols}{2}
}{%
    \end{multicols}
    \end{tcolorbox}
    \end{figure*}
}

\section{Case Study}
\label{sec:case_study}

\begin{figure}[h]
\centering
\small
\begin{tabular}{p{0.95\columnwidth}}
\toprule
\textbf{Question:} List the top 3 unique product categories by percentage of orders from California customers that were shipped late in 2023. \\
\midrule
\textbf{[Probe]} \\[2pt]
\textit{Probe 1:} \\
\texttt{SELECT DISTINCT state FROM customers LIMIT 5} \\
\textit{Result:} \texttt{[('CA',), ('NY',), ('TX',), ...]} \\ [1.5pt]
\textit{\textbf{Learned:}} \\ 
\textcolor{green!70!black}{State stored as 2-letter codes} \\[6pt]

\textit{Probe 2:} \\
\texttt{SELECT ship\_date, required\_date FROM orders LIMIT 3} \\
\textit{Result:} \texttt{[('2023-01-15', '2023-01-10'), ...]} \\ [1.5pt]
\textit{\textbf{Learned:}} \\
\textcolor{green!70!black}{Dates as YYYY-MM-DD} \\
\textcolor{green!70!black}{``Late'' means ship\_date > required\_date} \\
\midrule

\textbf{[Generate]} \\[2pt]
\texttt{SELECT p.category, CAST(SUM(CASE WHEN o.ship\_date > o.required\_date THEN 1 ELSE 0 END) AS REAL) * 100 / COUNT(*) FROM orders o JOIN customers c ON o.customer\_id = c.id JOIN products p ON o.product\_id = p.id WHERE c.state = 'CA' GROUP BY p.category} \\
\midrule

\textbf{[Verify]} \\[2pt]
\textbf{Extracted Constraints:} \\
\hspace{1em}\ding{51} Percentage calculation required \\
\hspace{1em}\ding{51} DISTINCT or GROUP BY for uniqueness \\
\hspace{1em}\textcolor{red!80!black}{\ding{55}} ``top 3'' requires ORDER BY ... LIMIT 3 $\rightarrow$ \textcolor{red!80!black}{\textbf{Violation}} \\
\hspace{1em}\textcolor{red!80!black}{\ding{55}} ``2023'' must appear in SQL $\rightarrow$ \textcolor{red!80!black}{\textbf{Violation}} \\
\midrule
\textbf{[Repair]} \\[2pt]
\texttt{SELECT p.category, CAST(SUM(...) AS REAL) * 100 / COUNT(*) \sethlcolor{green!15}\hl{AS late\_pct} FROM orders o JOIN customers c ... JOIN products p ... WHERE c.state = 'CA' \hl{AND strftime('\%Y', o.order\_date) = '2023'} GROUP BY p.category \hl{ORDER BY late\_pct DESC LIMIT 3}} \\
\midrule
\textbf{[Verify]} \hfill \textcolor{green!70!black}{\ding{51} All constraints satisfied} \\
\bottomrule
\end{tabular}
\caption{Case study showing how probing discovers value formats and verification catches multiple semantic errors (missing year filter, missing LIMIT).}
\label{fig:case_study}
\end{figure}

Figure~\ref{fig:case_study} demonstrates how probing and verification work together in {\tool}. The question asks for the top 3 product categories by late shipment percentage for California customers in 2023.

Through \emph{Probe}, the agent discovers that states are stored as 2-letter codes and dates follow the \texttt{YYYY-MM-DD} format, revealing that ``late'' means \texttt{ship\_date > required\_date}. Without probing, these value formats would be difficult to infer from schema alone.
\emph{Verify} extracts four constraints from the question: percentage calculation, uniqueness (via GROUP BY), top-3 limit, and year filter. The initial SQL satisfies two constraints but misses the ORDER BY ... LIMIT clause and the 2023 filter.
\emph{Repair} fixes both violations (highlighted in green), and a final verification confirms all constraints are satisfied.
This design aligns with broader work on agent design that improves reliability through structured interaction loops~\cite{all_fem,alloy,construct_ai,spa,atar}, reducing the burden of generating everything correctly in a single pass.

\begin{table}[h]
\centering
\small
\begin{tabular}{p{2.2cm} p{4.8cm}}
\toprule
\textbf{Constraint} & \textbf{Trigger Patterns} \\
\midrule
Distinctness & ``unique'', ``distinct'', ``different'', ``no duplicate'', ``deduplicate'' \\
\midrule
Top-K & ``top \textit{N}'', ``first \textit{N}'', ``bottom \textit{N}'', ``highest \textit{N}'', ``lowest \textit{N}'', ``best \textit{N}'', ``worst \textit{N}'' \\
\midrule
Ranking & ``rank'', ``ranking'', ``position'', ``placed'', ``standing'' \\
\midrule
Counting & ``how many'', ``count'', ``number of'', ``total number'', ``quantity of'' \\
\midrule
Percentage & ``percentage'', ``percent'', ``\%'', ``ratio'', ``rate'', ``proportion'', ``fraction of'' \\
\midrule
Summation & ``total'', ``sum'', ``overall'', ``combined'', ``aggregate'' \\
\midrule
Average & ``average'', ``mean'', ``avg'', ``on average'', ``typical'' \\
\midrule
Extreme & ``maximum'', ``minimum'', ``max'', ``min'', ``largest'', ``smallest'', ``most'', ``least'', ``highest'', ``lowest'' \\
\midrule
Temporal & ``latest'', ``earliest'', ``most recent'', ``newest'', ``oldest'', ``last'', ``first'' (with time context) \\
\midrule
Comparison & ``more than'', ``less than'', ``greater than'', ``fewer than'', ``at least'', ``at most'', ``no more than'', ``exceeds'' \\
\bottomrule
\end{tabular}
\caption{Constraint extraction rules mapping question patterns to semantic constraints. Patterns are matched case-insensitively against the question text.}
\label{tab:extraction_rules}
\end{table}

\begin{table}[h]
\centering
\small
\begin{tabular}{p{2.2cm} p{4.8cm}}
\toprule
\textbf{Constraint} & \textbf{SQL Verification Criteria} \\
\midrule
Distinctness & Presence of \texttt{DISTINCT} keyword in \texttt{SELECT}, or \texttt{GROUP BY} clause that implies uniqueness \\
\midrule
Top-K & Presence of \texttt{LIMIT \textit{N}} clause where \textit{N} matches the requested count; combined with \texttt{ORDER BY} for ``top'' queries \\
\midrule
Ranking & Presence of window function: \texttt{RANK()}, \texttt{DENSE\_RANK()}, or \texttt{ROW\_NUMBER()} with \texttt{OVER} clause \\
\midrule
Counting & Presence of \texttt{COUNT(*)} or \texttt{COUNT(column)} in \texttt{SELECT} clause \\
\midrule
Percentage & Presence of division operator (\texttt{/}) or multiplication by \texttt{100.0} in \texttt{SELECT}; often with \texttt{CAST} for float division \\
\midrule
Summation & Presence of \texttt{SUM(column)} aggregation function in \texttt{SELECT} clause \\
\midrule
Average & Presence of \texttt{AVG(column)} aggregation function in \texttt{SELECT} clause \\
\midrule
Extreme & Presence of \texttt{MAX()} or \texttt{MIN()} function; or \texttt{ORDER BY} with \texttt{LIMIT 1} pattern \\
\midrule
Temporal & \texttt{ORDER BY} clause on date/time column; direction (\texttt{ASC}/\texttt{DESC}) matches ``earliest''/``latest'' \\
\midrule
Comparison & \texttt{WHERE} or \texttt{HAVING} clause with comparison operators (\texttt{>}, \texttt{<}, \texttt{>=}, \texttt{<=}) matching the constraint \\
\bottomrule
\end{tabular}
\caption{Constraint verification rules. Each rule checks for specific SQL constructs that satisfy the semantic requirement.}
\label{tab:verification_rules}
\end{table}

\section{Additional Experiment Results}
\label{sec:appendix_results}

This section contains additional experimental results that complement the main paper. We provide: (1) baseline LLM performance by difficulty across all models, (2) cross-benchmark comparison with GPT-4.1-mini to complement Table~\ref{tab:benchmark_cost_tradeoff}, (3) performance breakdown by difficulty on Mini-Dev with GPT-4o to complement Table~\ref{tab:by_difficulty_stacked}, and (4) performance breakdown by difficulty for GPT-4.1-mini.

\section{Sensitivity Analysis on Maximum Probing Iterations}
\label{sec:sensitivity_k}

This section expands on the choice of the maximum probing budget $K$ in Section~\ref{sec:probing}. We set $K=5$ by default to match the maximum repair budget $M=5$ used in Section~\ref{sec:generation}, motivated by our observation that the agent issues only 2.65 probes per task on average (Section~\ref{sec:component_eval}).

To understand how performance and cost scale with $K$, we vary $K$ from 0 (no probing) to 7 on the BIRD benchmark using GPT-4o as the base model. Table~\ref{tab:sensitivity_k} reports the resulting execution accuracy and average output tokens per task. Performance improves rapidly from $K=0$ to $K=3$, after which it largely saturates: gains beyond $K=3$ are marginal (within 0.5 EX points), while output tokens continue to grow roughly linearly. This indicates diminishing returns once the agent has issued a few targeted probes, and supports our default choice of $K=5$ as a conservative upper bound that rarely binds in practice.

\begin{table}[h]
\centering
\small
\setlength{\tabcolsep}{6pt}
\renewcommand{\arraystretch}{1.15}
\begin{tabular}{lcc}
\toprule
\textbf{$K$} & \textbf{EX} & \textbf{Output Tokens} \\
\midrule
$K = 0$ (no probing) & 62.13 & 67 \\
$K = 1$ & 63.30 & 112 \\
$K = 2$ & 64.47 & 157 \\
$K = 3$ & 64.86 & 184 \\
$K = 4$ & 64.67 & 223 \\
$K = 5$ & 65.12 & 248 \\
$K = 6$ & 65.19 & 240 \\
$K = 7$ & 65.19 & 251 \\
\bottomrule
\end{tabular}
\caption{Sensitivity of execution accuracy (EX) and output tokens to the maximum probing budget $K$ on BIRD with GPT-4o.}
\label{tab:sensitivity_k}
\end{table}

\section{Execution Failures During Probing}
\label{sec:probe_failures}

This section expands on the discussion in Section~\ref{sec:component_eval} regarding how {\tool} handles execution failures during the probing phase. Probing queries are intentionally simple: each probe inspects a single aspect (e.g., the distinct values of one attribute or the format of a date column) and avoids complex joins or aggregations. As a result, they are far less prone to execution errors than full SQL synthesis.

Empirically, we measure that probe queries fail to execute only \textbf{0.03\%} of the time across all BIRD tasks. When a failure does occur, the agent treats the returned database error message as additional context and regenerates the probe in the next iteration of the loop in Algorithm~\ref{alg:framework}. Because each probe is independently re-issued under the same probing budget $K$, transient failures do not propagate to the final SQL generation step.

\section{Constraint Coverage Analysis}
\label{sec:constraint_coverage}

This section provides a headroom analysis that complements the error analysis in Section~\ref{sec:results}. Our rule-based extractor achieves 99.39\% precision (Table~\ref{tab:component_eval}) but is intentionally not designed to cover every possible semantic constraint. To quantify how many failures are attributable to constraints that the current rules do not extract, we conduct an additional analysis on all queries that {\tool} fails to solve correctly on BIRD with GPT-4o.

Specifically, for each failed case we reconstruct an oracle constraint list by extracting constraints from the \emph{ground-truth} SQL using the same constraint format as {\tool}, and then rerun the verify-and-repair loop with this oracle list (everything else, including the base model, probing context, and rule-based verifier, is held fixed). Under this perfect-extraction setting, \textbf{15.7\%} of previously failed cases are corrected. Given the high extraction precision of the existing rules (99.39\%, Table~\ref{tab:component_eval}), this improvement mainly stems from constraints that (i) are not covered by the current rule set and (ii) were also not successfully inferred by the LLM during refinement. This bounds the contribution of expanding the rule set or improving constraint extraction, and complements the precision-oriented design discussed in Section~\ref{sec:generation}.

\section{Prompts Used in {\tool}}
\label{sec:appendix_prompts}

We present the four prompts used in {\tool}: Probe, Generation, Repair, and Error Analysis.

\subsection{Probe Prompt}
\label{sec:prompt_probe}

The Probe prompt guides the LLM to generate probing SQL queries that reveal database content, value formats, and data distributions not apparent from schema alone. Given the question, evidence, schema, and results from prior probes, it outputs a JSON object containing a probe SQL query and discovered value mappings. This grounds the text-to-SQL task by understanding actual database content before generating the final query.

\subsection{Generation Prompt}
\label{sec:prompt_generation}

The Generation prompt produces the initial SQL query using enriched context from the probing phase. It takes the question, evidence, schema, value mappings, probe observations, and extracted constraints as input, and outputs a single SQL query that leverages the grounded understanding of database content.

\subsection{Repair Prompt}
\label{sec:prompt_repair}

The Repair prompt fixes SQL queries that violate constraints detected by the rule-based verifier. Given the original context, the faulty SQL, and specific violation messages from the verifier, it outputs a corrected SQL query. This enables iterative refinement by addressing specific constraint violations rather than regenerating from scratch.

\section{Error Analysis Prompt}
\label{sec:prompt_error_analysis}

The Error Analysis prompt is used for post-hoc analysis of failed predictions, classifying errors to understand model weaknesses. Given the question, evidence, schema, ground truth SQL, predicted SQL, and execution errors, it outputs a JSON object containing the error type, reasoning, and specific issue. This enables systematic categorization of errors for comparative analysis across methods.

\section{Probe Quality Evaluation Prompt}
\label{sec:probe_judge_prompt}

This section presents the prompt used for LLM-based evaluation of probe quality in Section~\ref{sec:results}. We use GPT-4o as the judge to assess each probe query along three dimensions: relevance to the question, whether it provides new insights beyond schema information, and redundancy with previous probes.

\section{LLM-based Verification Prompts}
\label{sec:llm_verify_prompts}

This section presents the prompts used in the ``LLM verify'' ablation variant described in Section~\ref{llm_verify}. In this variant, we replace the rule-based constraint extraction and verification with LLM-based approaches. While this approach is more flexible, it achieves lower accuracy and higher cost compared to rule-based verification.

\subsection{LLM-based Constraint Extraction}
\label{sec:prompt_llm_constraint}

The LLM-based constraint extraction prompt asks the model to analyze the question and extract semantic constraints that the SQL query must satisfy. Unlike the rule-based approach (Section~\ref{sec:rules_extract}), this relies on the LLM's understanding to identify requirements.

\subsection{LLM-based SQL Verification}
\label{sec:prompt_llm_verify}

The LLM-based verification prompt asks the model to verify whether a generated SQL query correctly satisfies all requirements from the original question. This replaces the rule-based verification that checks for specific SQL constructs.

\section{Constraint Extraction Rules}
\label{sec:rules_extract}

This section expands on the constraint extraction described in Section~\ref{sec:constraint_extraction}. Table~\ref{tab:extraction_rules} presents the complete set of pattern-matching rules for extracting semantic constraints from natural language questions. Each rule maps question patterns (keywords or phrases) to a constraint type that can be verified against the generated SQL.

\section{Constraint Verification Rules}
\label{sec:rules_verify}

This section expands on the violation detection described in Section~\ref{sec:generation}. Table~\ref{tab:verification_rules} presents the verification rules that check whether a generated SQL query satisfies each extracted constraint. Verification is performed by parsing the SQL and checking for the presence of required constructs.

We implement verification using SQL parsing with the \texttt{sqlparse} library. For each constraint type, we traverse the parsed AST to check for the required tokens or clauses. The verifier returns a list of violated constraints with descriptive error messages (e.g., ``Question asks for unique values but SQL lacks DISTINCT or GROUP BY'').

\begin{table*}[h]
\centering
\small

\setlength{\tabcolsep}{3.1pt}
\renewcommand{\arraystretch}{1.15}

\begin{tabular}{l c c p{2pt} c c p{2pt} c}
\toprule
\multirow{2}{*}{\textbf{Method}} &
\multicolumn{2}{c}{\textbf{BIRD}} & &
\multicolumn{2}{c}{\textbf{Mini-Dev}} & &
\multicolumn{1}{c}{\textbf{Spider}} \\
\cmidrule(lr){2-3}\cmidrule(lr){5-6}\cmidrule(lr){8-8}
& \metric{EX} & \metric{VES} &
& \metric{EX} & \metric{VES} &
& \metric{EX} \\
\midrule

GPT-4.1-mini
& 50.26 & 53.75 & & 49.40 & 54.86 & & 56.99 \\

\specialrule{0.1pt}{1.3pt}{1.3pt}

+ DAIL-SQL
& 53.80 & 62.71 & & 50.40 & 55.05 & & 58.77 \\

+ DIN-SQL
& 55.50 & 62.38 & & 53.80 & 58.22 & & 56.75 \\

+ MAC-SQL
& 58.30 & 63.97 & & 55.60 & 60.62 & & 62.77 \\

+ E-SQL
& 57.80 & 65.03 & & 54.80 & 60.95 & & 64.65 \\

+ TA-SQL
& 61.10 & 69.62 & & 60.20 & 66.11 & & 68.83 \\

+ XiYan-SQL
& 55.90 & 63.92 & & 51.60 & 58.99 & & 63.96 \\

+ TS-SQL
& 53.52 & 58.81 & & 50.20 & 54.87 & & 63.43 \\

\rowcolor{gray!12}
+ PV-SQL
& \textbf{63.62} & \textbf{72.78} & \cellcolor{gray!12}{}
& \textbf{60.20} & \textbf{67.15} & \cellcolor{gray!12}{}
& \textbf{75.55} \\
\bottomrule
\end{tabular}

\caption{Cross-benchmark comparison with GPT-4.1-mini as the base model.}
\label{tab:appendix_gpt41mini_benchmark}
\end{table*}

\begin{table*}[h]
\centering
\small

\setlength{\tabcolsep}{2.5pt}
\renewcommand{\arraystretch}{1.15}

\begin{tabular}{l cc cc cc}
\toprule
\multirow{2}{*}{\textbf{Method}} &
\multicolumn{2}{c}{\textbf{Simple}}
& \multicolumn{2}{c}{\textbf{Moderate}}
& \multicolumn{2}{c}{\textbf{Challenging}} \\
\cmidrule(lr){2-3}\cmidrule(lr){4-5}\cmidrule(lr){6-7}
& \metric{EX} & \metric{VES}
& \metric{EX} & \metric{VES}
& \metric{EX} & \metric{VES} \\
\midrule

GPT-4o
& 60.81 & 64.27 & 48.00 & 57.00 & 36.27 & 34.53 \\

\specialrule{0.1pt}{1.2pt}{1.2pt}

+ DAIL-SQL
& 66.89 & 68.13 & 47.60 & 48.24 & 31.37 & 32.12 \\
+ DIN-SQL
& 64.86 & 65.04 & 48.40 & 55.42 & 37.25 & 38.37 \\
+ MAC-SQL
& 72.30 & 82.81 & 55.20 & 62.60 & 43.14 & 43.34 \\
+ E-SQL
& 69.59 & 84.82 & 55.60 & 58.94 & 44.12 & 47.32 \\
+ TA-SQL
& 72.97 & 81.91 & 57.60 & 62.82 & 39.22 & 41.17 \\
+ XiYan-SQL
& 65.54 & 67.30 & 50.00 & 56.58 & 38.24 & 39.72 \\
+ TS-SQL
& 67.57 & 76.82 & 54.00 & 55.93 & 39.22 & 41.37 \\

\rowcolor{gray!12}
+ PV-SQL
& \textbf{76.35} & \textbf{87.28} & \textbf{62.80} & \textbf{76.21} & \textbf{48.04} & \textbf{52.40} \\

\bottomrule
\end{tabular}

\caption{Performance by difficulty on Mini-Dev with GPT-4o.}
\label{tab:appendix_gpt4o_minidev_difficulty}
\end{table*}

\begin{table*}[h]
\centering
\small

\setlength{\tabcolsep}{2.5pt}
\renewcommand{\arraystretch}{1.15}

\begin{tabular}{l cc cc cc}
\toprule
\multirow{2}{*}{\textbf{Method}} &
\multicolumn{2}{c}{\textbf{Simple}}
& \multicolumn{2}{c}{\textbf{Moderate}}
& \multicolumn{2}{c}{\textbf{Challenging}} \\
\cmidrule(lr){2-3}\cmidrule(lr){4-5}\cmidrule(lr){6-7}
& \metric{EX} & \metric{VES}
& \metric{EX} & \metric{VES}
& \metric{EX} & \metric{VES} \\
\midrule

GPT-4.1-mini
& 58.05 & 61.68 & 39.44 & 44.29 & 35.17 & 33.42 \\

\specialrule{0.1pt}{1.2pt}{1.2pt}

+ DAIL-SQL
& 61.30 & 74.23 & 44.60 & 47.74 & 35.90 & 37.20 \\

+ DIN-SQL
& 61.80 & 69.65 & 48.30 & 54.45 & 37.90 & 41.35 \\

+ MAC-SQL
& 65.30 & 73.68 & 47.60 & 50.84 & 48.30 & 44.11 \\

+ E-SQL
& 64.00 & 74.42 & 48.50 & 51.70 & 47.60 & 47.78 \\

+ TA-SQL
& 68.10 & 78.22 & 51.90 & 57.59 & 45.50 & 53.22 \\

+ XiYan-SQL
& 63.40 & 72.77 & 46.10 & 52.25 & 39.30 & 44.81 \\

+ TS-SQL
& 60.32 & 68.66 & 45.26 & 45.77 & 36.55 & 37.69 \\

\rowcolor{gray!12}
+ PV-SQL
& \textbf{69.73} & \textbf{79.21} & \textbf{56.47} & \textbf{66.46} & \textbf{47.59} & \textbf{51.95} \\

\bottomrule
\end{tabular}

\caption{Performance by difficulty on BIRD with GPT-4.1-mini.}
\label{tab:appendix_gpt41mini_difficulty}
\end{table*}

\begin{table*}[h]
\centering
\small

\setlength{\tabcolsep}{2.5pt}
\renewcommand{\arraystretch}{1.15}

\begin{tabular}{l cc cc cc}
\toprule
\multirow{2}{*}{\textbf{Method}} &
\multicolumn{2}{c}{\textbf{Simple}}
& \multicolumn{2}{c}{\textbf{Moderate}}
& \multicolumn{2}{c}{\textbf{Challenging}} \\
\cmidrule(lr){2-3}\cmidrule(lr){4-5}\cmidrule(lr){6-7}
& \metric{EX} & \metric{VES}
& \metric{EX} & \metric{VES}
& \metric{EX} & \metric{VES} \\
\midrule

GPT-4.1-mini
& 63.51 & 73.28 & 48.40 & 53.28 & 31.37 & 31.98 \\

\specialrule{0.1pt}{1.2pt}{1.2pt}

+ DAIL-SQL
& 67.60 & 78.02 & 47.60 & 50.47 & 32.40 & 32.96 \\

+ DIN-SQL
& 64.20 & 73.58 & 54.80 & 57.33 & 36.30 & 38.13 \\

+ MAC-SQL
& 70.90 & 78.36 & 52.40 & 58.34 & 41.20 & 40.47 \\

+ E-SQL
& 66.20 & 76.53 & 54.40 & 58.99 & 39.20 & 43.13 \\

+ TA-SQL
& 76.40 & 86.07 & 58.40 & 64.63 & 41.20 & 40.77 \\

+ XiYan-SQL
& 65.50 & 76.20 & 49.60 & 55.63 & 36.30 & 42.26 \\

+ TS-SQL
& 62.84 & 74.13 & 48.80 & 50.61 & 35.29 & 37.35 \\

\rowcolor{gray!12}
+ PV-SQL
& \textbf{73.65} & \textbf{86.52} & \textbf{59.20} & \textbf{64.04} & \textbf{43.14} & \textbf{46.68} \\

\bottomrule
\end{tabular}

\caption{Performance by difficulty on Mini-Dev with GPT-4.1-mini.}
\label{tab:appendix_gpt41mini_minidev}
\end{table*}

\begin{table*}[t]
\centering
\small
\setlength{\tabcolsep}{2.6pt}
\renewcommand{\arraystretch}{1.2}

\begin{tabular}{l cccccccccccc}
\toprule
\multirow{2}{*}{\textbf{Difficulty}} & \multicolumn{2}{c}{\textbf{GPT-4o}} & \multicolumn{2}{c}{\textbf{GPT-4.1-mini}} & \multicolumn{2}{c}{\textbf{GPT-OSS-20B}} & \multicolumn{2}{c}{\textbf{Gemma3-4B}} & \multicolumn{2}{c}{\textbf{Qwen3-4B}} & \multicolumn{2}{c}{\textbf{Qwen3-0.6B}} \\
\cmidrule(lr){2-3}\cmidrule(lr){4-5}\cmidrule(lr){6-7}\cmidrule(lr){8-9}\cmidrule(lr){10-11}\cmidrule(lr){12-13}
& \textbf{EX} & \textbf{VES} & \textbf{EX} & \textbf{VES} & \textbf{EX} & \textbf{VES} & \textbf{EX} & \textbf{VES} & \textbf{EX} & \textbf{VES} & \textbf{EX} & \textbf{VES} \\
\midrule

Simple & 65.00 & 70.06 & 60.80 & 67.88 & 63.57 & 66.16 & 44.22 & 45.58 & 44.97 & 49.86 & 18.70 & 22.03 \\

Moderate & 45.30 & 48.74 & 42.70 & 46.18 & 45.04 & 47.35 & 21.12 & 21.88 & 23.49 & 22.74 & 4.96 & 6.14 \\

Challenging & 38.60 & 40.35 & 33.10 & 34.41 & 40.00 & 40.07 & 15.86 & 16.36 & 23.45 & 21.90 & 3.45 & 4.02 \\

\specialrule{0.1pt}{1.2pt}{1.2pt}

All & 56.50 & 60.80 & 52.70 & 58.16 & 55.74 & 58.00 & 34.55 & 35.65 & 36.44 & 39.01 & 13.10 & 15.52 \\

\bottomrule
\end{tabular}

\caption{ZS-CoT performance by difficulty level across different LLMs on BIRD.}
\label{tab:appendix_baseline_bird}
\end{table*}

\begin{table*}[t]
\centering
\small
\setlength{\tabcolsep}{8pt}
\renewcommand{\arraystretch}{1.2}

\begin{tabular}{l cccc}
\toprule
\multirow{2}{*}{\textbf{Difficulty}} & \multicolumn{2}{c}{\textbf{GPT-4o}} & \multicolumn{2}{c}{\textbf{GPT-4.1-mini}} \\
\cmidrule(lr){2-3}\cmidrule(lr){4-5}
& \textbf{EX} & \textbf{VES} & \textbf{EX} & \textbf{VES} \\
\midrule

Simple & 60.81 & 64.27 & 66.90 & 79.65 \\

Moderate & 48.00 & 57.00 & 48.40 & 54.67 \\

Challenging & 36.27 & 34.53 & 31.40 & 32.99 \\

\specialrule{0.1pt}{1.2pt}{1.2pt}

All & 49.40 & 54.57 & 50.40 & 57.64 \\

\bottomrule
\end{tabular}

\caption{ZS-CoT performance by difficulty level on Mini-Dev..}
\label{tab:appendix_baseline_minidev}
\end{table*}

\begin{prompt}{Prompt 1: Probe}
\textcolor{sectioncolor}{\#\# Background}\\
You are helping to solve a text-to-SQL problem. Before writing the final SQL query, you can run exploratory ``probe'' queries on the database to understand its content. Probes help discover addition knowledge that are not apparent from the schema alone.\\[0.5em]
\textcolor{sectioncolor}{\#\# Task}\\
Analyze the question and schema, then decide whether to request a probe query or proceed to SQL generation.\\[0.5em]
\textcolor{sectioncolor}{\#\# Context}\\
Question: \{question\}\\
Evidence (hints provided with the question): \{evidence\}\\
Database Schema: \{schema\}\\
Prior Probes (queries you already ran and their results): \{prior\_probe\_results\}\\[0.5em]
\textcolor{sectioncolor}{\#\# Instructions}\\
- If you need more information, generate a probe query (e.g., SELECT DISTINCT column FROM table LIMIT 5).\\
- If you have enough information, set action to "done".\\
- Record any value mappings you discover (e.g., "California" maps to "CA" in the database).\\[0.5em]
\textcolor{sectioncolor}{\#\# Output Format (JSON)}\\
\{\\
\hspace*{1em}"action": "probe" | "done",\\
\hspace*{1em}"probe\_sql": "SELECT ...",\\
\hspace*{1em}"relevant\_columns": \{"table": ["col1", "col2"]\},\\
\hspace*{1em}"value\_mappings": \{"term\_in\_question": "exact\_db\_value"\}\\
\}
\end{prompt}

\begin{prompt}{Prompt 2: Generation}
\textcolor{sectioncolor}{\#\# Task}\\
Write a SQL query to answer the question based on the provided context.\\[0.5em]
\textcolor{sectioncolor}{\#\# Context}\\
Question: \{question\}\\
Evidence (hints provided with the question): \{evidence\}\\
Database Schema: \{schema\}\\[0.5em]
Probe Observations (results from exploratory queries run on the database, showing actual values and formats):\\
\{probe\_results\}\\[0.5em]
Extracted Constraints (requirements derived from the question, e.g., needs DISTINCT, needs LIMIT, needs COUNT):\\
\{constraints\}\\[0.5em]
\textcolor{sectioncolor}{\#\# Rules}\\
- Write a single SQL statement only.\\
- Return only what is asked (no extra columns).\\
- Follow evidence/hints strictly when provided.\\
- Use exact database values discovered from probes (e.g., use "CA" not "California" if probes showed state codes).\\[0.5em]
\textcolor{sectioncolor}{\#\# Output}\\
SQL query only.
\end{prompt}

\begin{prompt}{Prompt 3: Repair}
\textcolor{sectioncolor}{\#\# Background}\\
A SQL query was generated but failed verification. The verifier checks for constraint violations (e.g., missing DISTINCT when the question asks for unique values, missing LIMIT for top-k queries) and execution errors (e.g., invalid column names, syntax errors).\\[0.5em]
\textcolor{sectioncolor}{\#\# Task}\\
Fix the SQL query to resolve the detected violations.\\[0.5em]
\textcolor{sectioncolor}{\#\# Context}\\
Question: \{question\}\\
Evidence (hints provided with the question): \{evidence\}\\
Database Schema: \{schema\}\\[0.5em]
Original SQL (the query that failed verification):\\
\{original\_sql\}\\[0.5em]
\textcolor{red}{\#\# Violations Detected (errors found by the verifier):}\\
\{violation\_messages\}\\[0.5em]
\textcolor{sectioncolor}{\#\# Instructions}\\
- Carefully address each violation listed above.\\
- Output the corrected SQL only (no explanation).
\end{prompt}

\begin{prompt}{Prompt 4: Error Analysis}
\textcolor{sectioncolor}{\#\# Background}\\
You are analyzing why a text-to-SQL model failed to generate the correct query. By comparing the predicted SQL with the ground truth, you will classify the root cause of the error to help understand model weaknesses.\\[0.5em]
\textcolor{sectioncolor}{\#\# Task}\\
Analyze the failure and classify it into one error category.\\[0.5em]
\textcolor{sectioncolor}{\#\# Context}\\
Question: \{question\}\\
Evidence (hints provided with the question): \{evidence\}\\
Database Schema: \{schema\}\\
Ground Truth SQL (the correct answer): \{gold\_sql\}\\
Predicted SQL (what the model generated): \{predicted\_sql\}\\
Execution Error (if the predicted SQL failed to run): \{exec\_error\}\\[0.5em]
\textcolor{sectioncolor}{\#\# Error Types}\\
Classify into exactly one category:\\[0.3em]
1. \textbf{DATABASE\_MISINTERPRETATION}:\\
Failed to understand database content, schema, or relationships.\\
Examples: wrong table/column, misunderstood foreign keys or data formats.\\[0.3em]
2. \textbf{QUESTION\_MISINTERPRETATION}:\\
Misinterpreted the question.\\
Examples: wrong filter conditions, misunderstood aggregation requirements.\\[0.3em]
3. \textbf{SQL\_SYNTHESIS\_FAILURE}:\\
Understood context correctly but failed to generate correct SQL.\\
Examples: wrong JOIN syntax, missing clauses, syntax errors.\\[0.5em]
\textcolor{sectioncolor}{\#\# Output Format (JSON)}\\
\{\\
\hspace*{1em}"error\_type": "...",\\
\hspace*{1em}"reasoning": "...",\\
\hspace*{1em}"specific\_issue": "..."\\
\}
\end{prompt}

\begin{prompt}{Prompt 5: Probe Quality Evaluation}
\textcolor{sectioncolor}{\#\# Background}\\
You are evaluating the quality of probe queries for text-to-SQL grounding. Probes are exploratory SQL queries that help understand database content before generating the final query.\\[0.5em]
\textcolor{sectioncolor}{\#\# Task}\\
For each probe query, assess three aspects:\\[0.3em]
1. \textbf{RELEVANCE}: Is this probe relevant to answering the question? (yes/no)\\
2. \textbf{NEW\_INSIGHT}: Does this probe provide information not available from schema alone? (yes/no)\\
\hspace{1em}- Schema-only info: table/column names, data types, foreign keys\\
\hspace{1em}- New insights: actual data values, data formats, value distributions, NULL patterns\\
3. \textbf{REDUNDANT}: Does this probe duplicate information from previous probes? (yes/no)\\[0.5em]
\textcolor{sectioncolor}{\#\# Context}\\
Question: \{question\}\\
Evidence: \{evidence\}\\
Schema: \{schema\}\\
Probes to evaluate: \{probes\}\\[0.5em]
\textcolor{sectioncolor}{\#\# Output Format (JSON)}\\
\{\\
\hspace*{1em}"evaluations": [\\
\hspace*{2em}\{"probe\_index": 0, "relevant": true/false,\\
\hspace*{2.5em}"new\_insight": true/false, "redundant": true/false,\\
\hspace*{2.5em}"reasoning": "..."\},\\
\hspace*{2em}...\\
\hspace*{1em}]\\
\}
\end{prompt}

\begin{prompt}{Prompt 6: LLM-based Constraint Extraction}
\textcolor{sectioncolor}{\#\# Task}\\
You are a SQL requirements analyst. Extract semantic constraints from the natural language question and evidence.\\[0.5em]
\textcolor{sectioncolor}{\#\# Context}\\
Question: \{question\}\\
Evidence: \{evidence\}\\[0.5em]
\textcolor{sectioncolor}{\#\# Instructions}\\
Focus on identifying:\\
- DISTINCT requirements (unique, different, distinct values)\\
- Aggregation needs (count, sum, avg, max, min)\\
- Ordering/ranking requirements (top N, highest, lowest, oldest, newest)\\
- Percentage/ratio calculations\\
- Comparison operators (greater than, less than, equal to)\\
- Grouping requirements\\
- NULL handling needs\\
- Any specific value filtering mentioned\\[0.5em]
\textcolor{sectioncolor}{\#\# Output Format (JSON)}\\
\{\\
\hspace*{1em}"constraints": [\\
\hspace*{2em}\{\\
\hspace*{3em}"type": "distinct|limit|aggregation|...",\\
\hspace*{3em}"description": "human-readable description",\\
\hspace*{3em}"sql\_hint": "what SQL construct should be used"\\
\hspace*{2em}\}\\
\hspace*{1em}],\\
\hspace*{1em}"output\_requirements": \{\\
\hspace*{2em}"expected\_columns": ["col1", "col2"],\\
\hspace*{2em}"expected\_type": "single\_value|list|count|..."\\
\hspace*{1em}\}\\
\}
\end{prompt}

\begin{prompt}{Prompt 7: LLM-based SQL Verification}
\textcolor{sectioncolor}{\#\# Task}\\
You are a SQL verification expert. Verify if the generated SQL query correctly satisfies all requirements from the original question.\\[0.5em]
\textcolor{sectioncolor}{\#\# Context}\\
Question: \{question\}\\
Evidence: \{evidence\}\\
Extracted Constraints: \{constraints\}\\
Schema: \{schema\}\\
Generated SQL: \{sql\}\\[0.5em]
\textcolor{sectioncolor}{\#\# Verification Checklist}\\
Verify ALL aspects:\\
1. \textbf{Structural validity}: Is it a single valid SELECT/WITH statement?\\
2. \textbf{Semantic correctness}: Does the SQL return what the question asks?\\
3. \textbf{Filter correctness}: Are all filters/conditions applied?\\
4. \textbf{Aggregation correctness}: Is aggregation correct (COUNT vs COUNT(DISTINCT))?\\
5. \textbf{Ordering correctness}: Is ordering/limit correct for ``top N'' queries?\\
6. \textbf{JOIN correctness}: Are JOINs correct and complete?\\
7. \textbf{Output format}: Is the output format correct?\\[0.5em]
\textcolor{sectioncolor}{\#\# Instructions}\\
Be thorough but avoid false positives. Only flag issues that are clearly problems.\\[0.5em]
\textcolor{sectioncolor}{\#\# Output Format (JSON)}\\
\{\\
\hspace*{1em}"is\_valid": true/false,\\
\hspace*{1em}"issues": [\\
\hspace*{2em}\{\\
\hspace*{3em}"severity": "error|warning",\\
\hspace*{3em}"category": "syntax|semantic|missing\_constraint|...",\\
\hspace*{3em}"description": "detailed description of the issue",\\
\hspace*{3em}"suggestion": "how to fix it"\\
\hspace*{2em}\}\\
\hspace*{1em}]\\
\}
\end{prompt}

\end{document}